\documentclass[runningheads]{llncs}

\usepackage[mobile]{eccv}

\usepackage{eccvabbrv}

\titlerunning{Be-Your-Outpainter}

\authorrunning{Fu-Yun Wang et al.}

\usepackage{graphicx}
\usepackage{booktabs}
\usepackage{multirow} 
\usepackage{algorithm}
\usepackage{algpseudocode}

\usepackage[accsupp]{axessibility}

\usepackage[pagebackref,breaklinks,colorlinks,citecolor=eccvblue]{hyperref}
\usepackage{wrapfig}
\usepackage{minitoc}
\usepackage{enumitem}
\usepackage{marvosym}

\usepackage{orcidlink}

\usepackage{amsmath,amsfonts,bm,amssymb}
\usepackage[most]{tcolorbox}

\def\1{\bm{1}}

\def\rmI{{\mathbf{I}}}

\def\rmW{{\mathbf{W}}}

\def\vmu{{\bm{\mu}}}
\def\vtheta{{\bm{\theta}}}

\def\vepsilon{{\bm{\epsilon}}}

\def\vm{{\bm{m}}}

\def\vp{{\bm{p}}}

\def\vv{{\bm{v}}}

\def\vx{{\bm{x}}}

\DeclareMathAlphabet{\mathsfit}{\encodingdefault}{\sfdefault}{m}{sl}
\SetMathAlphabet{\mathsfit}{bold}{\encodingdefault}{\sfdefault}{bx}{n}

\def\gL{{\mathcal{L}}}

\def\gN{{\mathcal{N}}}

\def\sR{{\mathbb{R}}}

\usepackage[most]{tcolorbox}

\newtcolorbox{mblock}[1]
{
  colframe     = gray,
  coltitle     = black,
  colbacktitle = lightgray!50!white,
  colback      = white,
  title        = #1,
  fonttitle    = \bfseries,
  arc          = 0mm,
  left         = 2pt,
  right        = 2pt,
}

\newtcolorbox{rblock}[1]
{
  colframe     = green,
  coltitle     = gray,
  colbacktitle = lightgray!50!white,
  colback      = white,
  title        = \ifthenelse{\isempty{#1}}
                  {Remark}
                  {Remark: #1},
  fonttitle    = \bfseries,
  arc          = 0mm,
  left         = 2pt,
  right        = 2pt,
}

\newcommand{\ours}{MOTIA\xspace}

\begin{document}

\title{Be-Your-Outpainter: Mastering Video Outpainting through Input-Specific Adaptation}

\author{Fu-Yun Wang\textsuperscript{1*} \hspace{-3mm} \quad Xiaoshi Wu\textsuperscript{1*} \hspace{-3mm} \quad Zhaoyang Huang\textsuperscript{2 } \hspace{-3mm} \\ \quad Xiaoyu Shi\textsuperscript{1} \hspace{-3mm}\quad Dazhong Shen\textsuperscript{3} \hspace{-3mm}  \quad Guanglu Song\textsuperscript{4}  \quad Yu Liu\textsuperscript{3} \hspace{-2mm} \quad Hongsheng Li\textsuperscript{1 \Letter}\\ 
}

\institute{
\textsuperscript{1}MMLab, CUHK
\quad \textsuperscript{2}Avolution AI \quad 
\textsuperscript{3}Shanghai AI Lab \quad 
\textsuperscript{4}SenseTime Research\\
\email{\{fywang@link, hsli@ee\}.cuhk.edu.hk}
}

\maketitle

{
    \begin{center}
    \textbf{\url{https://be-your-outpainter.github.io/}}
    \end{center}
}

\begin{abstract}

Video outpainting is a challenging task, aiming at generating video content outside the viewport of the input video while maintaining inter-frame and intra-frame consistency. Existing methods fall short in either generation quality or flexibility. We introduce \ours{} (\textbf{M}astering Video \textbf{O}utpainting \textbf{T}hrough \textbf{I}nput-Specific \textbf{A}daptation), a diffusion-based pipeline that leverages both the intrinsic data-specific patterns of the source video and the image/video generative prior for effective outpainting. \ours{} comprises two main phases: input-specific adaptation and pattern-aware outpainting. The input-specific adaptation phase involves conducting efficient and effective pseudo outpainting learning on the single-shot source video. This process encourages the model to identify and learn patterns within the source video, as well as bridging the gap between standard generative processes and outpainting. The subsequent phase, pattern-aware outpainting, is dedicated to the generalization of these learned patterns to generate outpainting outcomes. Additional strategies including spatial-aware insertion and noise travel are proposed to better leverage the diffusion model's generative prior and the acquired video patterns from source videos. Extensive evaluations underscore \ours{}'s superiority, outperforming existing state-of-the-art methods in widely recognized benchmarks. Notably, these advancements are achieved without necessitating extensive, task-specific tuning. 
\end{abstract}

\section{Introduction}
\label{sec:intro}
\begin{figure*}[t!]
    \centering
    \makebox[\textwidth]{\includegraphics[width=1.15\textwidth]{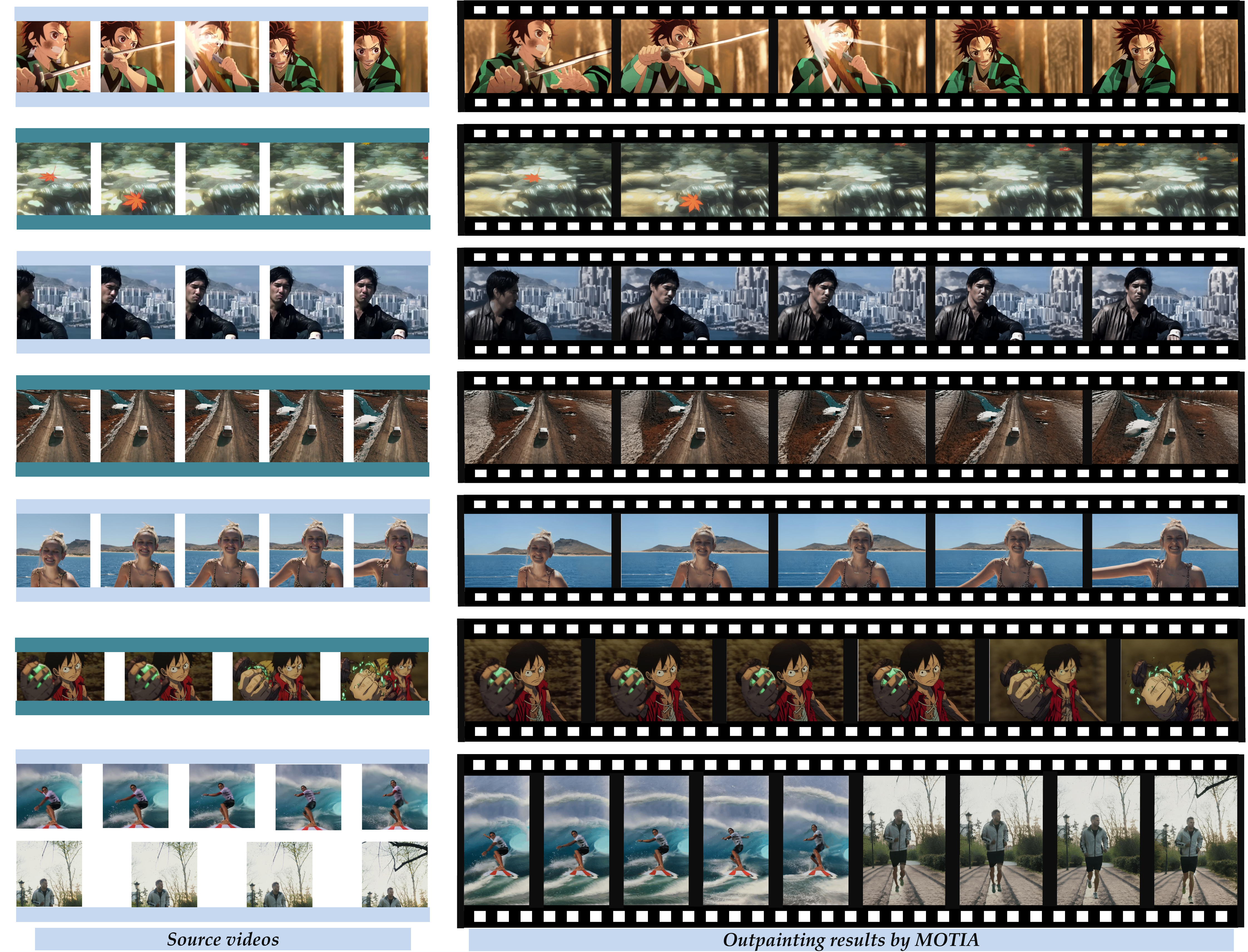}}
    \caption{\small MOTIA is a high-quality flexible video outpainting pipeline, leveraging the intrinsic data-specific patterns of source videos and image/video generative prior for state-of-the-art performance. Quantitative metric improvement of MOTIA is significant~(Table~\ref{tab:quant}).}
    \label{fig:teaser}
\newpage
\end{figure*}

Video outpainting aims to expand the visual content out of the spatial boundaries of videos, which has important real-world applications~\cite{dehan,fan2023hierarchical,barnes2009patchmatch}. For instance, in practice, videos are usually recorded with a fixed aspect ratio, such as in movies or short clips. This becomes an issue when viewing these videos on smartphones, which often have varying aspect ratios, resulting in unsightly black bars on the screen that detract from the viewing experience. Proper ways for video outpainting are crucial in solving this issue. By expanding the visual content beyond the original frame, it adapts the video to fit various screen sizes seamlessly. This process ensures that the audience enjoys a full-screen experience without any compromise in visual integrity.  However, the challenges associated with video outpainting are significant. It requires not just the expansion of each frame's content but also the preservation of temporal~(inter-frame) and spatial~(intra-frame) consistency across the video.

\begin{wrapfigure}{r}{0.6\textwidth}
     \centering
\includegraphics[width=\linewidth]{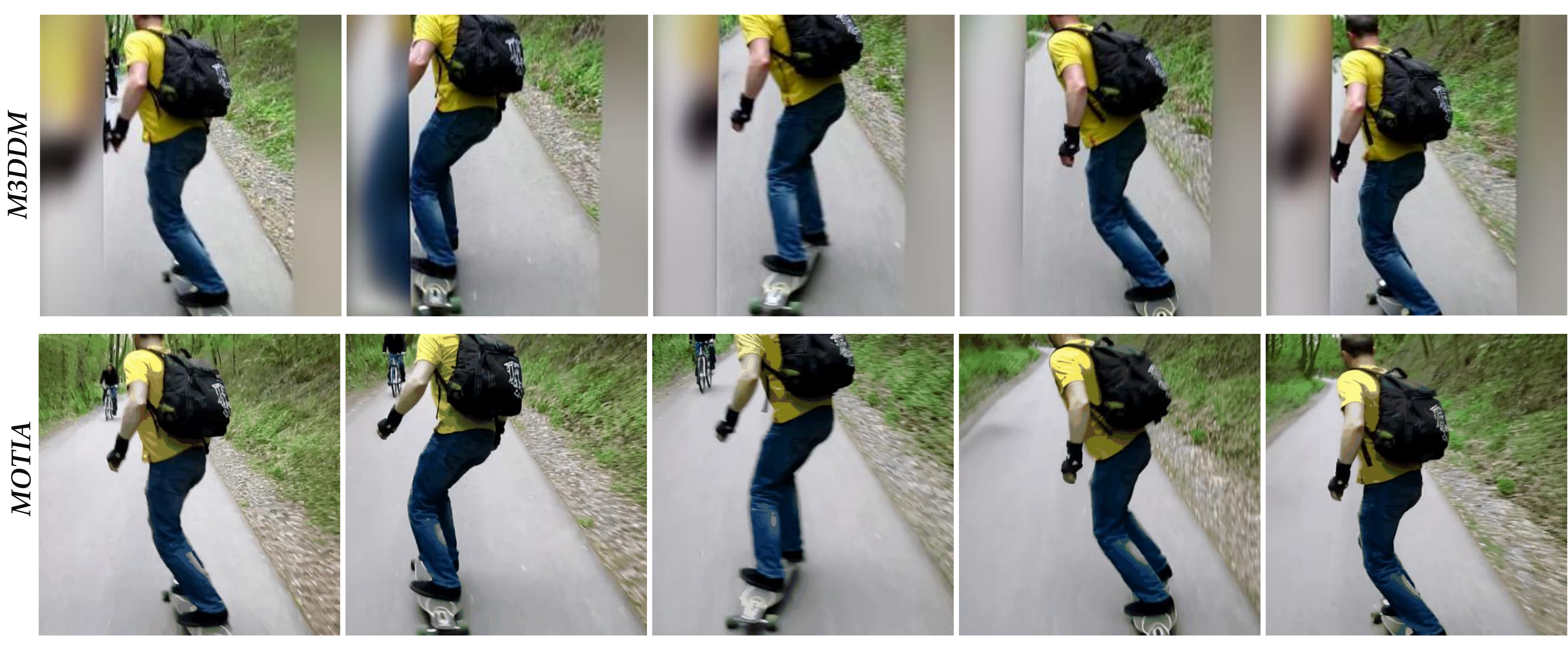}
    \caption{\textbf{Failure example of previous methods.} Many previous methods including the intensively trained models on video outpainting still might suffer from generation failure, that the model simply generates blurred corners. \ours{} never encounters this failure.}
    \label{fig:drawbacks}
\end{wrapfigure}

Currently, there are two primary approaches to video outpainting. The first employs optical flows and specialized warping techniques to extend video frames,  involving complex computations and carefully tailored hyperparameters to ensure the added content remains consistent~\cite{dehan,gao2020flow}. However, their results are far from satisfactory, suffering from blurred content. The other type of approach in video outpainting revolves around training specialized models tailored for video inpainting and outpainting with extensive datasets~\cite{fan2023hierarchical,yu2023magvit}. However, they have two notable limitations: 1) An obvious drawback of these models is their dependency on the types of masks and the resolutions of videos they can handle, which significantly constrains their versatility and effectiveness in real-world applications, as they may not be adequately equipped to deal with the diverse range of video formats and resolutions commonly encountered in practical scenarios. 2) The other drawback is their inability to out-domain video outpainting, even intensively trained on massive video data. Fig.~\ref{fig:drawbacks} shows a failure example of most advanced previous work~\cite{fan2023hierarchical} that the model faces complete outpainting failure, with only blurred corners. We show that a crucial reason behind this is that the model fails at capturing the intrinsic data-specific patterns from out-domain source~(input) videos.

In this work, we propose \textbf{\ours{}}: \textbf{M}astering Video \textbf{O}utpainting \textbf{T}hrough \textbf{I}nput-Specific \textbf{A}daptation, a diffusion-based method for open-domain video outpainting with arbitrary types of mask, arbitrary video resolutions and lengths, and arbitrary styles.  At the heart of \ours{} is treating the source video itself as a rich source of information~\cite{nikankin2022sinfusion,shaham2019singan}, which contains key motion and content patterns~(intrinsic data-specific patterns) necessary for effective outpainting. We demonstrate that the patterns learned from the source video, combined with the generative capabilities of diffusion models, can achieve surprisingly great outpainting performance. 

\begin{figure}[!t]
    \centering   \makebox[\textwidth]{\includegraphics[width=1.15\linewidth]{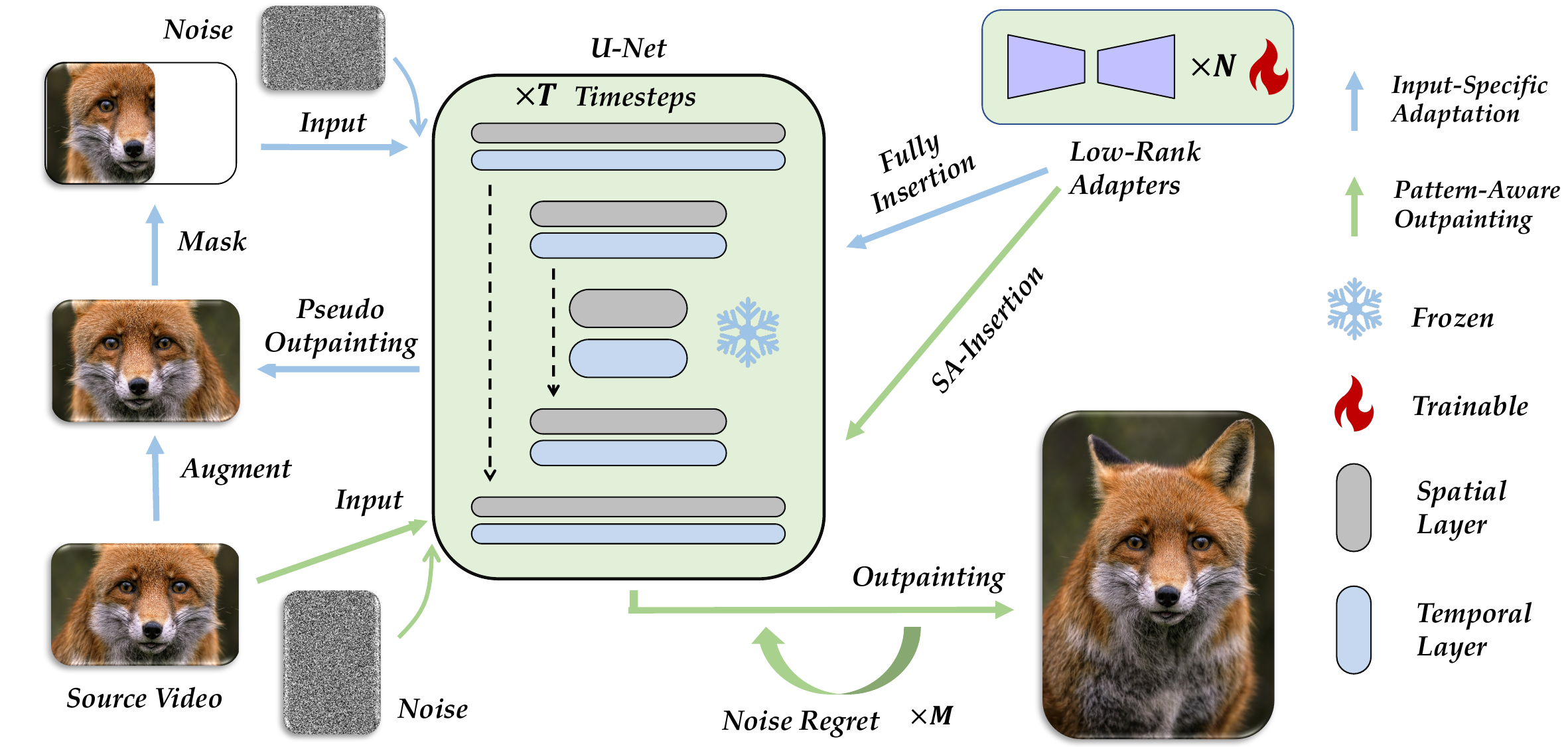}}
    \caption{\textbf{Workflow of \ours.} Blue lines represent the workflow of input-specific adaptation, and green lines represent the workflow of pattern-aware outpainting.}
    \label{fig:overview}
\end{figure}

Fig.~\ref{fig:overview} illustrates the workflow of \ours{}. \ours{} consists of two stages: input-specific adaptation and pattern-aware outpainting. During the input-specific adaptation stage, we conduct pseudo video outpainting learning on the source video~(videos to be outpainted) itself. Specifically, at each iteration, we heuristically add random masks to the source video and prompt the base diffusion model to recover the masked regions by learning to denoise the video corrupted by white noise, relying on the extracted information from unmasked regions. This process not only allows the model to capture essential data-specific patterns from the source video but also narrows the gap between standard generation and outpainting. We insert trainable lightweight adapters into the diffusion model for tuning to keep the efficiency and stability. In the pattern-aware outpainting stage, we combine the learned patterns from the source video and the generation prior of the diffusion model for effective outpainting. To better leverage the generation ability of the pretrained diffusion model and the learned pattern from the source video, we propose spatial-aware insertion~(SA-Insertion) of the tuned adapters for outpainting. Specifically, the insertion weights of adapters gradually decay as the spatial position of features away from the known regions. In this way, the outpainting of pixels near the known regions is more influenced by the learned patterns, while the outpainting of pixels far from the known regions relies more on the original generative prior of diffusion model.  To further mitigate potential denoising conflicts and enhance the knowledge transfer between known regions and unknown regions, we incorporate noise regret that we add noise and denoise periodically at early inference steps, which works for more harmonious outpainting results.

Extensively quantitative and qualitative experiments verify the effectiveness of our proposed method. \ours{} overcomes many limitations of previous methods and outperforms the state-of-the-art intensively trained outpainting method in standard widely used benchmarks. In summary, our contribution is three-fold: 1) We show that the data-specific patterns of source videos are crucial for effective outpainting, which is neglected by previous work. 2) We introduce an adaptation strategy to effectively capture the data-specific patterns and then propose novel strategies to better leverage the captured patterns and pretrained image/video generative prior for better results. 3) Vast experiments verify that our performance in video outpainting is great, significantly outperforming previous state-of-the-art methods in both quantitative metrics and user studies.

\section{Related Works}
\label{sec:relatedwork}
In this section, we discuss related diffusion models and outpainting methods.

\noindent \textbf{Diffusion models.}
Diffusion models~(a.k.a., score-based models)~\cite{song2020denoising,ddpm,nichol2021glide,rombach2022high,ho2022imagen} have gained increasing attention due to its amazing ability to generate highly-detailed images. Current successful  video diffusion models~\cite{blattmann2023align,singer2022make,ho2022imagen,voleti2022mcvd} are generally built upon the extension of image diffusion models through inserting temporal layers. They are either trained with image-video joint tuning~\cite{singer2022make,ho2022video} or trained with spatial weights frozen~\cite{blattmann2023align} to mitigate the negative influence of the poor captions and visual quality of video data. 

\noindent \textbf{Ooutpainting methods.}
 Video outpainting is largely built upon the advancements in image outpainting, where techniques ranged from patch-based methods (\eg, PatchMatch~\cite{barnes2009patchmatch}) to more recent deep learning approaches like GANs~\cite{xu2018attngan,arora2021singan-gif}. Diffusion models~\cite{lugmayr2022repaint,avrahami2022blended}, benefiting from the learned priors on synthesis tasks and the process of iterative refinement, achieve state-of-the-art performance on image outpainting tasks. The research focusing on video outpainting is comparatively few. Previous works typically apply optical flow for outpainting, which warps content from adjacent frames to the outside corners, but their results are far from satisfactory. Recently, M3DDM~\cite{fan2023hierarchical} trained a large 3D diffusion models with specially designed architecture for outpainting on massive video data, achieving much better results compared to previous methods, showcasing the huge potential of diffusion models on video outpainting. However, as we claimed, they have two main limitations: 1) The inflexibility for mask types and video resolutions. They can only outpaint video with resolution $256 \times 256$ with square type of masking. 2) Inability for out-domain video outpainting. As shown in Fig.~\ref{fig:drawbacks}, they encounter outpainting failure when processing out domain videos even intensively trained on massive video data.

\section{Preliminaries}\label{sec:preliminary}
\noindent\textbf{Diffusion models} \cite{ddpm} add noise to data through a Markov chain process. Given initial data \( \vx_0 \sim q(\vx_0) \), this process is formulated as:
\begin{equation}
    q(\vx_{1:T}|\vx_0) = \prod_{t=1}^T q(\vx_t|\vx_{t-1}), \quad        q(\vx_t|\vx_{t-1}) = \mathcal{N}(\vx_t|\sqrt{\alpha_t} \vx_{t-1}, \beta_t \mathbf{I}),
\end{equation}
where \( \beta_t \) is the noise schedule and \( \alpha_t = 1 - \beta_t \). The data reconstruction, or denoising process, is accomplished by the reverse transition modeled by \( p_\theta(\vx_{t-1}|\vx_t) \):
\begin{equation}
    q(\vx_{t-1}|\vx_t, \vx_0) = \mathcal{N}(\vx_{t-1}; \tilde{\vmu}_t(\vx_t, \vx_0), \tilde{\beta}_t \mathbf{I}),
\end{equation}
with \(\tilde{\vmu}_t(\vx_t, \vx_0) = \frac{1}{\sqrt{\alpha_t}} \vx_t - \frac{1-\alpha_t}{\sqrt{1-\bar{\alpha}_t} \sqrt{\alpha_t}} \vepsilon\), \( \bar{\alpha}_t = \prod_{s=1}^t \alpha_s \), \( \tilde{\beta}_t = \frac{1 - \bar{\alpha}_{t-1}}{1 - \bar{\alpha}_t} \beta_t \), and \( \vepsilon \) is the noise added to \( \vx_0 \) to obtain \( \vx_t \). 

\noindent\textbf{Diffusion-based outpainting}
aims to predict missing pixels at the corners of the masked region with the pre-trained diffusion models. We denote the ground truth as $\vx$, mask as $\vm$, known region as $(\boldsymbol{1} - \vm) \odot \vx$, and unknown region as   $\vm \odot \vx$. At each reverse step in the denoising process, we modify the known regions by incorporating the intermediate noisy state of the source data from the corresponding timestep in the forward diffusion process (which adds noise), provided that this maintains the correct distribution of correspondences. Specifically, each reverse step can be denoted as the following formulas:
\begin{equation}\label{eq:diffusion-outpainting}
    \begin{split}
        \vx_{t-1}^{\text {known}}  \sim \mathcal{N}\left(\sqrt{\bar{\alpha}_t} \vx_0,\left(1-\bar{\alpha}_t\right) \mathbf{I}\right), \quad
\vx_{t-1}^{\text {unknown}} \sim \mathcal{N}\left(\vmu_\theta\left(\vx_t, t\right), \Sigma_\theta\left(x_t, t\right)\right),
    \end{split}
\end{equation}
\begin{equation}\label{eq:diffusion-outpainting2}
    \vx_{t-1}  =\vm \odot \vx_{t-1}^{\text {known }}+(\boldsymbol{1}-\vm) \odot \vx_{t-1}^{\text {unknown}} \, ,
\end{equation}
where the $\vx_{t-1}^{\text{known}}$ is padded to the target resolution before the masked merging.

\begin{figure*}[!t]
    \centering
\makebox[\textwidth]{\includegraphics[width=1.15\textwidth]{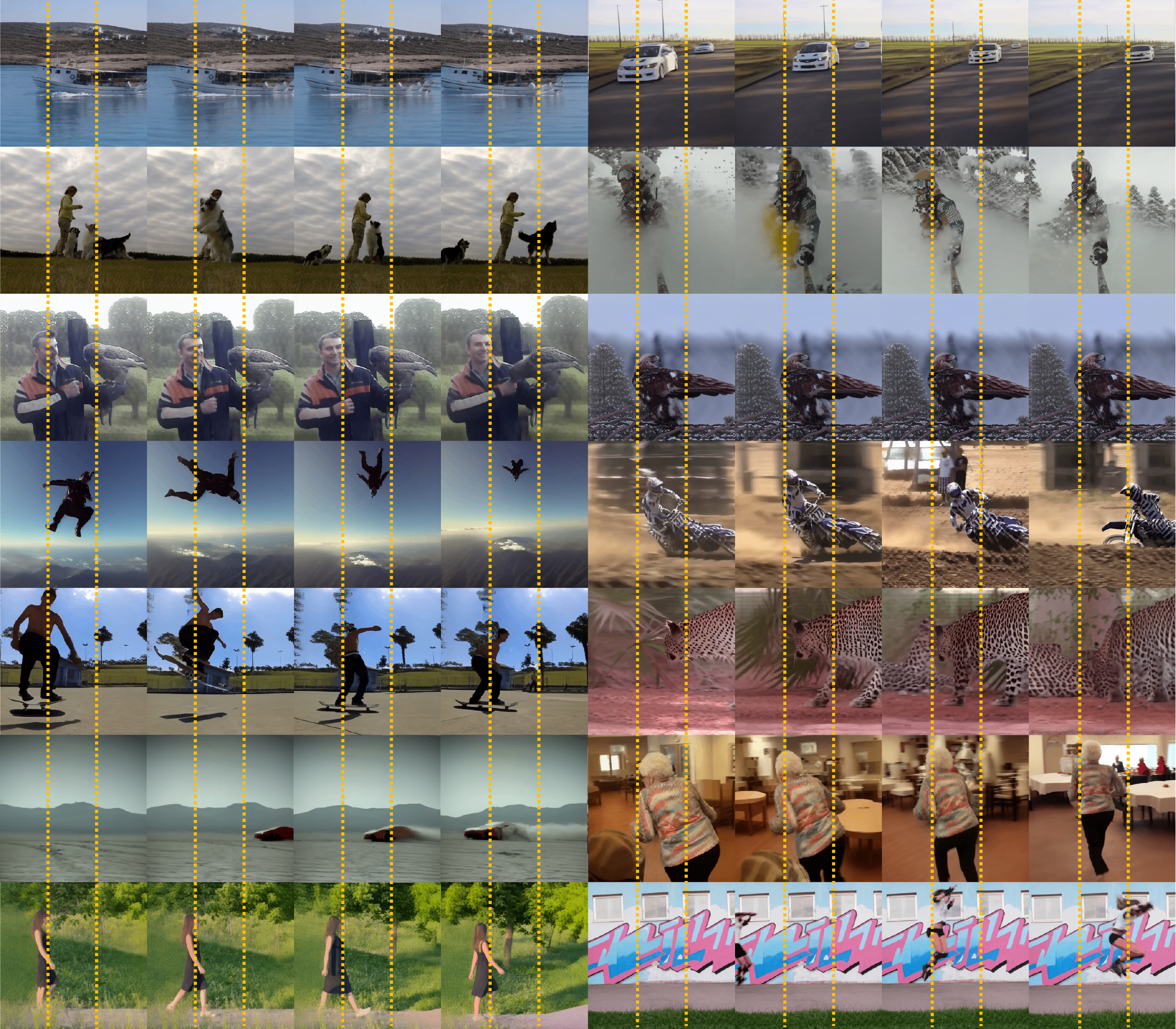}}
    \caption{\textbf{Sample results of quantitative experiments.} All videos are outpainted with a horizontal mask ratio of 0.66. Contents outside the yellow lines are outpainted by \ours{}.}
    \label{fig:qualitative_quantitative}
\end{figure*}

\section{Methodology}

This section presents \ours{}, a method demonstrating exceptional performance in video outpainting tasks. We begin by defining the concept of video outpainting and describing the foundational model in Section~\ref{sec:promblemformulation}. and Section~\ref{sec:networkexpansion}. Subsequently, we delve into the specifics of input-specific adaptation and pattern-aware outpainting in Sections~\ref{sec:intrinsicadaptation} and~\ref{sec:extrinsicrendering}, respectively. Moreover,  we show that our approach has great promise in extending its application to long video outpainting, which will be explored in Section~\ref{sec:longvideooutpainting}.

\subsection{Problem Formulation}\label{sec:promblemformulation}
 For a video represented as $\vv \in \sR^{t \times h \times w \times d}$, where \( t \) denotes the number of frames, \( h \) denotes frame height, \( w \) denotes frame width, and \( d \) denotes channel depth. Video outpainting model \( f(\vv) \) is designed to generate a spatially expanded video \( \vv' \in \sR^{t \times h' \times w' \times d} \).  This process not only increases the spatial dimensions ($h' > h$, $w' > w$), but also requires to ensure continuity and harmony between the newly expanded regions and the known regions. The transformation maintains the known regions unchanged, with $f(\vv)$ acting as an identity in these regions.
\subsection{Network Expansion}\label{sec:networkexpansion}

\noindent \textbf{Network inflation.} 
 \ours{} leverages the pre-trained text-to-image (T2I) model, Stable Diffusion. In line with previous video editing techniques~\cite{wu2022tune}, we transform 2D convolutions into pseudo 3D convolutions and adapt 2D group normalizations into 3D group normalizations to process video latent features. Specifically, the $3\times3$ kernels in convolutions are replaced by $1\times3\times3$ kernels, maintaining identical weights. Group normalizations are executed across both temporal and spatial dimensions, meaning that all 3D features within the same group are normalized simultaneously, followed by scaling and shifting. 

\noindent \textbf{Masked video as conditional input.} Additionally, we incorporate a ControlNet~\cite{zhang2023adding},  initially trained for image inpainting, to manage additional mask inputs.  Apart from noise input, ControlNet can also process masked videos to extract effective information for more controllable denoising. In these masked videos, known regions have pixel values ranging from $0$ to $1$, while values of masked regions are set to $-1$. 

\noindent \textbf{Temporal consistency prior.}
To infuse the model with temporal consistency priors, we integrate temporal modules pre-trained on text-to-video (T2V) tasks.  Note that although \ours{} relies on pre-trained video diffusion modules, applying these pre-trained temporal modules directly for video outpainting yields rather bad results, significantly inferior to all baseline methods~(Table.~\ref{tab:ablation}). However, when equipped with our proposed \ours{}, the model demonstrates superior performance even in comparison to models specifically designed and trained for video outpainting, underscoring the efficacy of \ours{}.

\subsection{Input-Specific Adaptation}\label{sec:intrinsicadaptation}
The input-specific adaptation phase is crucial in our video outpainting method, aiming to tailor the model for the specific challenges of outpainting. This phase involves training on the source video with a pseudo-outpainting task, importantly, enabling the model to learn intrinsic content and motion patterns~(data-specific patterns) within the source video as well as narrowing the gap between the standard generation process and outpainting.

\noindent\textbf{Video augmentation.} Initially, we augment the source video. Transformations like identity transformation, random flipping, cropping, and resizing can be employed. This step can potentially help the model better learn and adapt to diverse changes in video content. For longer video outpainting, as we will discuss later, instead of taking it as a whole, we randomly sample short video clips from it to reduce the cost of the adaptation phase. 

\noindent\textbf{Video masking.} We then add random masks to the video. We adopt a heuristic approach that uniformly samples edge boundaries of 4 sides within given limits. The area enclosed by these boundaries is considered the known region, while the rest is the unknown region. This masked video serves as the conditional input for the ControlNet, simulating the distribution of known and unknown areas in actual outpainting scenarios.

\noindent\textbf{Video noising.} Additionally, we apply noise to the video following the DDPM~\cite{ddpm} by randomly selecting diffusion timesteps. This noisy video serves as an input for both the ControlNet and the Stable Diffusion model, training the model to adapt to various noise conditions.

\noindent\textbf{Optimization.} Finally, we optimize the model. To ensure efficiency, low-rank adapters are inserted into the layers of the diffusion model. We optimize only the parameters of these adapters while keeping the other parameters frozen. The loss function is 
\begin{equation}
\gL = \left\| \vepsilon - \hat{\vepsilon}_{\bar{\vtheta}_l, \bar{\vtheta}_c, \vtheta_a}(\vv_{\text{noisy}}, \vv_{\text{masked}}, t)\right\|_2 \, ,    
\end{equation}
where \( t \) represents the timestep in the process, \( \vepsilon \) is the added noise, \( \vv_{\text{noisy}} \) refers to the video perturbed by $\vepsilon$, and \( \vv_{\text{masked}} \) denotes the masked video. The parameters \( \vtheta_l \), \( \vtheta_c \), and \( \vtheta_a \) correspond to the Diffusion Model, ControlNet, and adapters, respectively.  The bar over these parameters indicates they are frozen during the optimization. This optimization process, including the steps of augmentation, masking, and noising, is repeated to update the lightweight adapters to capture the data-specific patterns from the source video.

\begin{figure}[t]
    \centering
    \begin{minipage}[t]{0.48\textwidth}
        \centering
        \includegraphics[width=\textwidth]{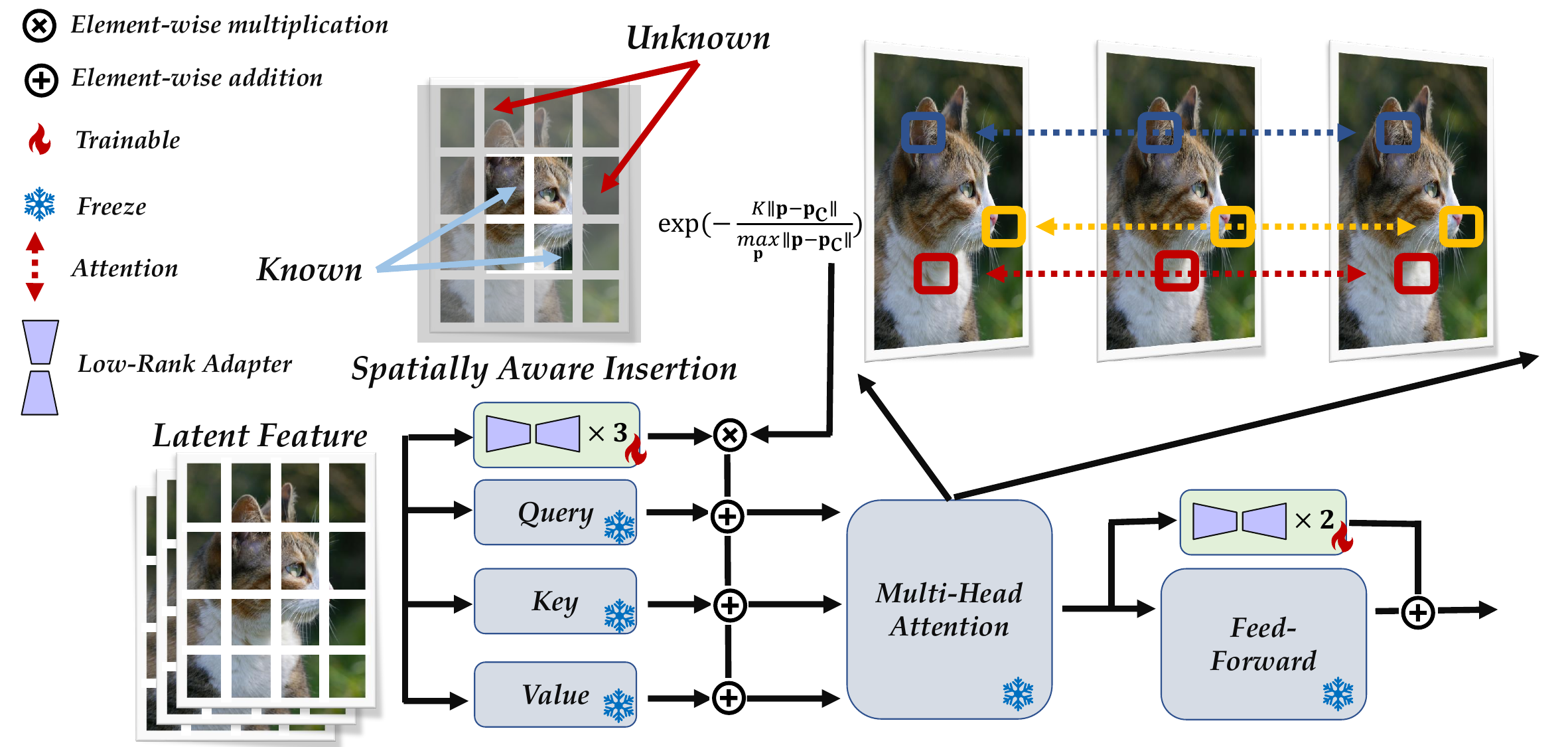}
        \caption{\textbf{Spatial-aware insertion} scales the insertion weights of adapters for better leveraging of learned patterns and generative prior.}
        \label{fig:SAInsetion}
    \end{minipage}
    \hfill
    \begin{minipage}[t]{0.48\textwidth}
        \centering
        \includegraphics[width=\textwidth]{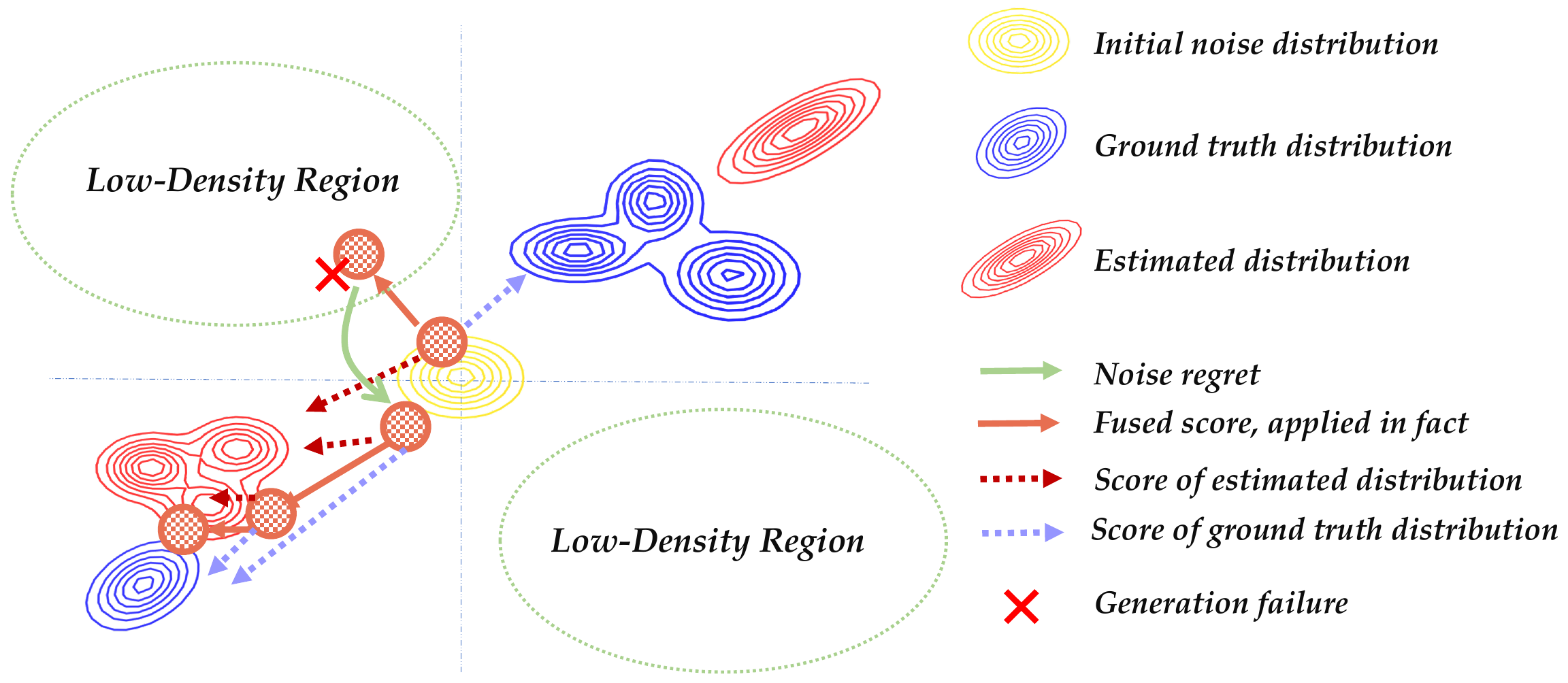}
        \caption{\textbf{Noise regret} fixes possible generation failure/degradation caused by score conflicts.}
        \label{fig:noiseregret}
    \end{minipage}
    \vspace{-4mm}
\end{figure}
\subsection{Pattern-Aware Outpainting}\label{sec:extrinsicrendering}

Following the initial phase of input-specific adaptation, our model shows promising results in video outpainting using basic pipelines as outlined in Eq.~\ref{eq:diffusion-outpainting} and Eq.~\ref{eq:diffusion-outpainting2}, achieving commendable quality. However, we here introduce additional inference strategies that can be combined to better leverage the learned data-specific patterns from the input-specific adaptation phase for better outpainting results. We call the outpainting process that incorporates these strategies pattern-aware outpainting.

\noindent\textbf{Spatial-aware insertion.}
 It is important to acknowledge that in the input-specific adaptation phase, the model is fine-tuned through learning outpainting within the source video. However, at the outpainting phase, the model is expected to treat the entire source video as known regions and then fill the unknown regions at edges~(\ie, generating a video with a larger viewport and resolution). This specificity may lead to a noticeable training-inference gap during outpainting, potentially affecting the outpainting quality. To balance the fine-tuned patterns with the diffusion model's inherent generative prior, we introduce the concept of spatial-aware insertion~(SA-Insertion) of adapters as shown in Fig.~\ref{fig:SAInsetion}. The adaptation involves adjusting the insertion weight of tuned low-rank adapters based on the feature's spatial position. We increase insertion weight near known areas to utilize the learned patterns while decreasing it in farther regions to rely more on the original generative capacity of the diffusion model. To be specific,
\begin{equation}
\rmW^\top_{\text{adapted}} \vx_{\vp} = \rmW^\top\vx_{\vp} + \alpha(\vp) \left(\rmW_{\text{up}} \rmW_{\text{down}}\right)^\top\vx_{\vp}.
\end{equation}
Here, $\vp$ signifies the spatial position of $\vx$, $\rmW\in \mathbb R^{d_\text{in}\times d_\text{out}}$ denotes the linear transformation in layers of diffusion model, $\rmW_\text{down}\in \mathbb R^{d_\text{in}\times r}$ and $\rmW_\text{up}\in \mathbb R^{r\times d_\text{out}}$ are the linear components of the adapter with rank $r\ll \min(d_\text{in},d_\text{out})$. The function $\alpha(\vp)$ is defined as:
\begin{equation}
\alpha(\vp) =
\exp (-\frac{K\|\vp - \vp_c\|}{\max_{\bar\vp} \| \bar\vp - \vp_c\|}) ,
\end{equation}
where $K$ is a constant for controlling decay speed, and $\vp_c$ represents the nearest side of the known region to $\vp$. 

\noindent \textbf{Noise regret.}
In the context of Eq.~\ref{eq:diffusion-outpainting}, merging noisy states from known and unknown regions in video outpainting tasks poses a technical problem. This process, similar to sampling from two different vectors, can disrupt the denoising direction. As depicted in Fig.~\ref{fig:noiseregret}, the estimated denoising direction initially points downwards to the left, in contrast to the true direction heading towards the top-right. This leads to a merged trajectory directed to a less dense top-left region, potentially resulting in generation failures (see Fig.~\ref{fig:drawbacks}), even in well-trained models. Given the significant impact of early steps on the generation's structure, later denoising may not rectify these initial discrepancies. Inspired by DDPM-based image inpainting methods~\cite{lugmayr2022repaint,rout2023theoretical}, we propose to re-propagate the noisy state into a noisier state by adding noise when denoising and then provide the model a second chance for re-denoising. This helps integrate known region data more effectively and reduces denoising direction conflicts. In detail, after obtaining $\vv_{t}$ during denoising, we conduct
\begin{equation}
\vv_{t+L} = \sqrt{\Pi_{i=t+1}^{t+L}\alpha_{i}}\vv_{t} + \sqrt{1-\Pi_{i=t+1}^{t+L}\alpha_{i}}\vepsilon, ,
\end{equation}
where $\alpha_{i}=1-\beta_{i}$ and $ \vepsilon \sim \gN(\boldsymbol{0},\rmI)$. Then we restart the denoising process. We repeat this progress for $M$ times. We only conduct it in the early denoising steps.
\subsection{Extension to Long Video Outpainting}\label{sec:longvideooutpainting}
We show that our method can be easily extended for long video outpainting. Specifically, for the stage of input-specific adaptation, instead of taking the long video as a whole for adaptation~(Direct adaptation on long videos is costly and does not align with the video generation prior of the pretrained modules), we randomly sample short video clips from the long video for tuning to learn global patterns without requiring more GPU memory cost. For the stage of pattern-aware outpainting, we split the long video into short video clips with temporal overlapping~(\ie, some frames are shared by different short video clips), and then conduct temporal co-denoising following Gen-L~\cite{wang2023gen}. Specifically, the denoising result for $j^{th}$ frame of the long video at timestep $t$ is approximated by the weighted sum of all the corresponding frames in short video clips that contain it,
\begin{equation}
    \boldsymbol{v}_{t-1, j}=\frac{\sum_{i \in \mathcal{I}^j}\left(\left(W_{i, j^*}\right)^2 \otimes \boldsymbol{v}_{t-1, j^*}^i\right)}{\sum_{i \in \mathcal{I}^j}\left(W_{i, j^*}^2\right)^2} \, ,
\end{equation}
where $\otimes$ denotes element-wise multiplication,$\vv_{t-1,j}$ denotes the noisy state of the $j^{th}$ frame at timestep $t$, $\vv_{t-1,j^*}^{i}$ is the noisy state of $j^{th}$ frame predicted with only information from the $i^{th}$ video clip at timestep $t$, $W_{i,j^*}$ is the per-pixel weight, which is as $\boldsymbol{1}$ as default.

\section{Experiments}

\subsection{Experimental Setup}

\noindent \textbf{Benchmarks.} To verify the effectiveness of \ours{}, we conduct evaluations on DAVIS~\cite{davis} and YouTube-VOS~\cite{vos}, which are widely used benchmarks for video outpainting.  Following M3DDM~\cite{fan2023hierarchical}, we compare the results of different methods in the horizontal direction, using mask ratios of 0.25 and 0.66.

\noindent \textbf{Evaluation metrics.}
Our evaluation approach utilizes four well-established metrics: Peak Signal to Noise Ratio~(PSNR), Structural Similarity Index Measure~(SSIM)~\cite{ssim}, Learned Perceptual Image Patch Similarity~(LPIPS)~\cite{lpips}, and Frechet Video Distance~(FVD)~\cite{fvd}. For assessing PSNR, SSIM, and FVD, the generated videos are converted into frames within a normalized value range of $[0, 1]$. LPIPS is evaluated over a range of $[-1, 1]$. About the FVD metric, we adopt a uniform frame sampling, with $16$ frames per video for evaluation following M3DDM.

\noindent \textbf{Compared methods.}
The comparative analysis includes the following methods: 1) 
\textbf{VideoOutpainting}~\cite{dehan}: Dehan et al.~\cite{dehan} propose to tackle video outpainting by bifurcating foreground and background components. It conducts separate flow estimation and background prediction and then fuses these to generate a cohesive output. 2) \textbf{SDM}~\cite{fan2023hierarchical}: SDM considers the initial and terminal frames of a sequence as conditional inputs, merged with the context at the initial network layer. It is trained on video datasets including WebVid~\cite{bain2021frozen} and e-commerce~\cite{fan2023hierarchical}. 3) \textbf{M3DDM}~\cite{fan2023hierarchical}: M3DDM is an innovative pipeline for video outpainting. It adopts a masking technique allowing the original source video as masked conditions. Moreover, it uses global-frame features for cross-attention mechanisms, allowing the model to achieve global and long-range information transfer. It is intensively trained on vast video data, including WebVid and e-commerce, with a specialized architecture design for video outpainting. In this way, SDM could be viewed as a pared-down version of M3DDM,  yet it is similarly intensively trained.

\begin{figure*}[!t]
    \centering
    \vspace{-2mm}
\makebox[\textwidth]{\includegraphics[width=1.15\textwidth]{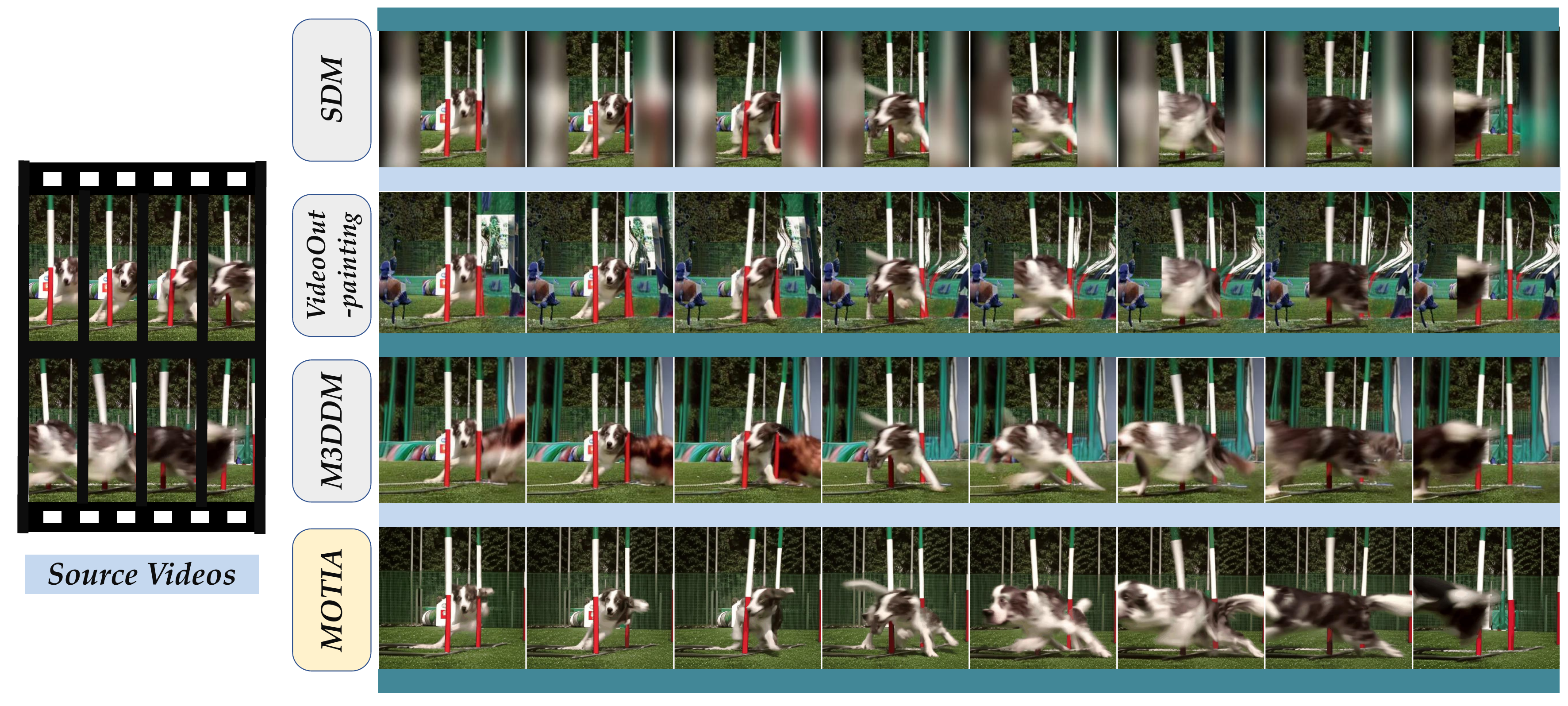}}
    \vspace{-4mm}
    \caption{\textbf{Qualitative comparison.} Other methods outpainting the source video with a mask ratio of 0.6. MOTIA outpainting the source video with a larger mask ratio of 0.66 while achieving obviously better outpainting results.}
    \label{fig:qualitative}
    \vspace{-4mm}
\end{figure*}

\noindent \textbf{Implementation details.}
Our method is built upon Stable Diffusion v1-5. We add the ControlNet pretrained on image inpainting to enable the model to accept additional masked image inputs. The temporal modules are initialized with the weights from pretrained motion modules~\cite{guo2023animatediff} to obtain additional motion priors. The motion modules are naive transformer blocks trained with solely text-to-video tasks on WebVid. For the input-specific adaptation, the low-rank adapters are trained using the Adam optimizer. We set the learning rate to $10^{-4}$, and set the weight decay to $10^{-2}$. The LoRA rank and $\alpha_\text{lora}$ are set to 16 and, 8, respectively. The number of training steps is set to 1000. We do not apply augmentation for simplicity. For both mask ratios of 0.66 and 0.25, we simply apply the same random mask strategy, which uniformly crops a square in the middle as the known regions. For the pattern-aware outpainting, the diffusion steps are set to 25 and the classifier-free guidance~(CFG) scale is set to 7.5 and we only apply CFG at the first 15 inference steps. When adding noise regret to further improve the results, we set jump length $L=3$, and repeat time $M=4$. We only apply noise regret in the first half inference steps. Note that our method is built upon LDM, which requires text-conditional inputs. For a fair comparison and to remove the influence of the choice of text prompt, we apply Blip~\cite{li2022blip} to select the prompt automatically. We observe dozens of prompt mistakes but do not modify them to avoid man-made influence.

\subsection{Qualitative Comparison}
Fig.~\ref{fig:qualitative} showcases a qualitative comparison of \ours{} against other methods. Outpainting a narrow video into a square format. \ours{} employs a mask ratio of 0.66, surpassing the 0.6 ratio utilized by other methods, and demonstrates superior performance even with this higher mask ratio. The SDM method only manages to blur the extremities of the video's background, egregiously overlooking the primary subject and resulting in the outpainting failure as previously highlighted in Fig.~\ref{fig:drawbacks}. Dehan's approach effectively outpaints the background but utterly fails to address the foreground, leading to notable distortions. In contrast, the M3DDM method adeptly handles both subject and background integration but is marred by considerable deviations in subject characteristics, such as incorrect brown coloration in the dog's fur across several frames. Our method stands out by achieving optimal results, ensuring a harmonious and consistent outpainting of both the foreground and background.
\begin{table*}[t]
\centering
\caption{\textbf{Quantitative comparison} of video outpainting methods on DAVIS and YouTube-VOS datasets. $\uparrow$ means `better when higher', and $\downarrow$ indicates `better when lower'.
}
\vspace{-1mm}
\resizebox{0.96\textwidth}{!}{%
\begin{tabular}{lccccccccc}
\hline
\multirow{2}{*}{Method} & \multicolumn{4}{c}{DAVIS}    & & \multicolumn{4}{c}{YouTube-VOS}   \\ \cline{2-5} \cline{7-10} 
 & PSNR $\uparrow$ & SSIM $\uparrow$ & LPIPS $\downarrow$ & FVD $\downarrow$ & & PSNR $\uparrow$ & SSIM $\uparrow$& LPIPS $\downarrow$ & FVD $\downarrow$ \\ \hline
VideoOutpainting~\cite{dehan} & 17.96 & 0.6272 & 0.2331&  363.1 & &  18.25 & 0.7195 & 0.2278 & 149.7 \\
SDM~\cite{fan2023hierarchical}  & 20.02 & 0.7078 &  0.2165 & 334.6 & & 19.91 &  0.7277 & 0.2001 &  94.81 \\
M3DDM~\cite{fan2023hierarchical} &  20.26 &  0.7082 &  0.2026 & 300.0 &  & 20.20 & 0.7312 & 0.1854 &  66.62 \\ 
\ours{} & \textbf{20.36} & \textbf{0.7578}  & \textbf{0.1595} & \textbf{286.3} &  & \textbf{20.25} & \textbf{0.7636} & \textbf{0.1727} & \textbf{58.99}\\ \hline
\end{tabular}%
}
\label{tab:quant}
\end{table*}

\subsection{Quantitative Comparison}
Table~\ref{tab:quant} summarizes the evaluation metrics of our method compared to other approaches. Our method achieves comparable results to the best method in PSNR. It shows significant improvements in video quality~(SSIM), perceptual metric~(LPIPS), and distribution similarity ~(FVD). Specifically, our SSIM, LPIPS, and FVD metrics show improvements of 7.00\%, 21.27\%, and 4.57\% respectively on the DAVIS dataset, and 4.43\%, 6.85\%, and 11.45\% on the YouTube-VOS dataset compared to the best-performing method.
\subsection{Ablation Study}
\begin{figure}[t]
    \centering
\makebox[\textwidth]{\includegraphics[width=1.15\linewidth]{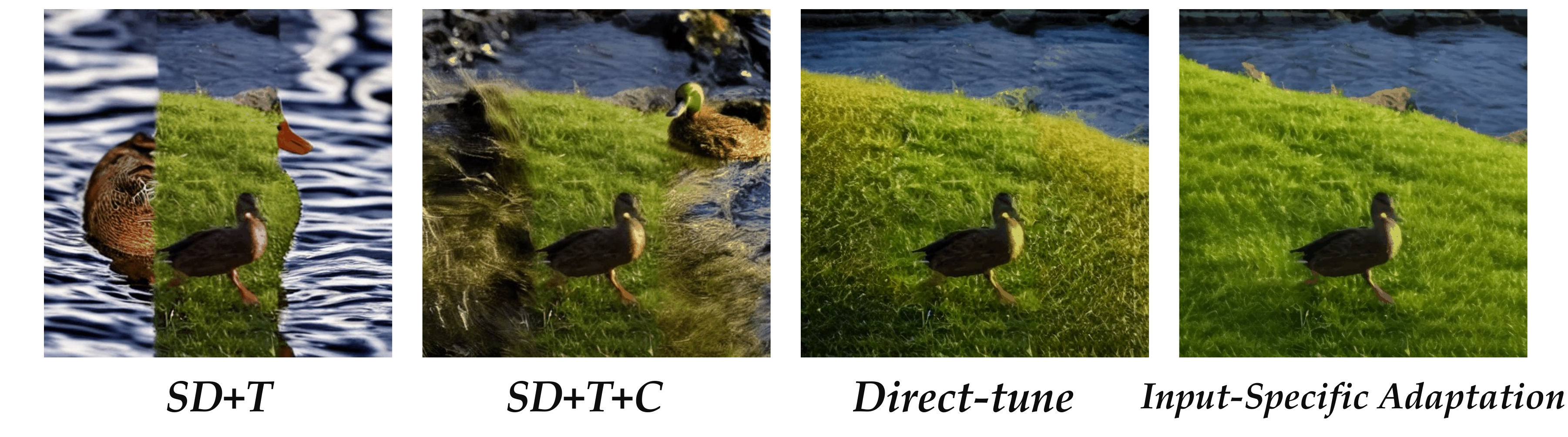}}
    \caption{\textbf{Visual examples} of ablation study on the proposed input-specific adaptation.}
    \label{fig:ablation1}
\end{figure}
\begin{figure}[t]
    \centering
\makebox[\textwidth]{\includegraphics[width=1.15\linewidth]{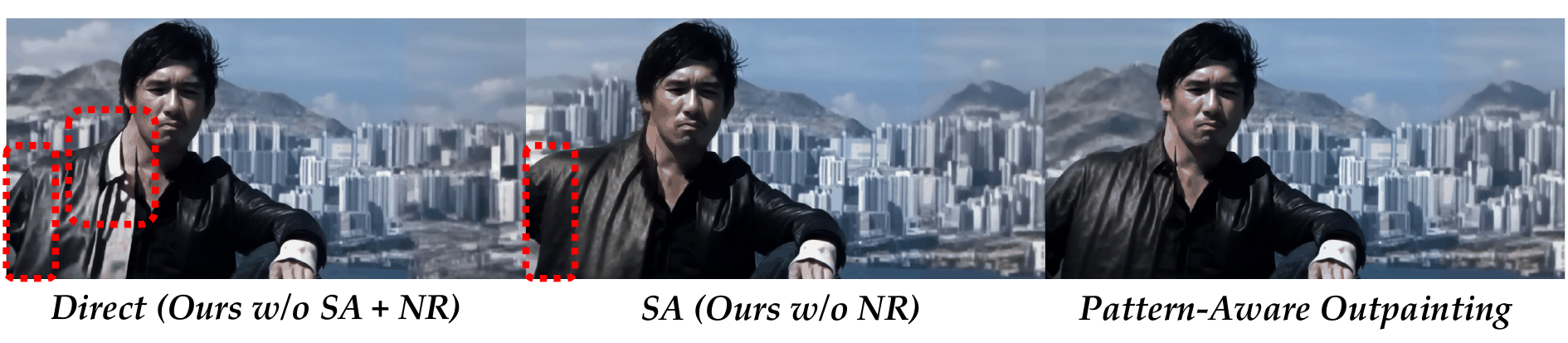}}
    \caption{\textbf{Visual examples} of ablation study on pattern-aware outpainting.}
    \label{fig:ablation2}
\end{figure}
\noindent \textbf{Ablation study on input-specific adaptation.}
We conducted the ablation study on input-specific adaptation with the DAVIS dataset to verify its effectiveness, as shown in Fig.~\ref{fig:ablation1} and Table~\ref{tab:ablation2}. ``SD+T'' represents the result of directly combining the temporal module with Stable Diffusion, which led to a complete outpainting failure. ``SD+T+C'' indicates the additional use of ControlNet, resulting in similarly poor outcomes. ``Direct-tune'' refers to the approach of directly fitting the original video without outpainting training; in this case, we observed a very noticeable color discrepancy between the outpainted and known areas. In contrast, our method achieved the best results, ensuring consistency in both the visual and temporal aspects. The metrics shown in Table~\ref{tab:ablation2} also support this observation, with \ours{} significantly outperforming the other baselines.

\noindent\textbf{Ablation study on pattern-aware outpainting.}
Table~\ref{tab:ablation} summarizes our ablation experiments for the pattern-aware outpainting part. We conducted extensive validation on the YouTube-VOS dataset. ``{Direct}'' refers to performing outpainting according to Eq.~\ref{eq:diffusion-outpainting} directly after input-specific adaptation. ``{SA}'' denotes spatially-aware insertion, and ``{SA+NR}'' indicates the combined use of spatially-aware insertion and noise regret. The experimental results demonstrate that each of our components effectively enhances performance. Specifically, Combining both SA-Insertion and Noise regret, the PSNR, SSIM, LPIPS, and FVD metrics show improvements of 2.69\%, 0.90\%,  3.95\%, and 11.32\% respectively than directly applying Eq.~\ref{eq:diffusion-outpainting}. Fig.~\ref{fig:ablation2} presents the visual examples of ablation study on our proposed pattern-aware outpainting part. When removing NR, it might fail to align the texture colors or produce unreasonable details (\eg, arms in the middle of
Fig.~\ref{fig:ablation2}). When further removing SA, it could potentially generate unrealistic results caused by the overfitting to the target video (\eg, the white collar on the left of Fig.~\ref{fig:ablation2}). Note that even though the FVD degrades in a very slight manner, all the other metrics increase and we qualitatively find it to be helpful for improving results.  
\begin{table}[t]
\vspace{-1mm}
\centering
\begin{minipage}[t]{0.45\textwidth}
\centering
\caption{\textbf{Ablation study} on input-specific adaptation. }
\vspace{-4mm}
\label{tab:ablation2}
\begin{tabular}{@{}lcccc@{}}
\toprule
Component    & PSNR$\uparrow$ & SSIM$\uparrow$ & LPIPS$\downarrow$ & FVD $\downarrow$\\ \midrule
SD+T      &    15.59   &    0.6640    &    0.2979   &   672.7     \\
SD+T+C      &    16.81    &  0.6961    &    0.2338    & 515.4       \\
Direct-tune     &     19.42   &    0.7375   &   0.1784     &  312.1      \\ 
\ours{}   &     20.36   &    0.7578    &   0.1595      &  286.3     \\ \bottomrule
\end{tabular}
\end{minipage}
\hfill
\begin{minipage}[t]{0.48\textwidth}
\centering
\caption{\textbf{Ablation study} on the proposed pattern-aware outpainting.}
\label{tab:ablation}
\vspace{-2mm}
\begin{tabular}{@{}lcccc@{}}
\toprule
Component    & PSNR$\uparrow$ & SSIM$\uparrow$ & LPIPS$\downarrow$ & FVD $\downarrow$\\ \midrule
Direct      &    19.72    &    0.7568    &    0.1798    &   66.52     \\
SA      &    19.97    &  0.7608      &    0.1752    & 58.40       \\
SA+NR     &     20.25   &    0.7636    &   0.1727      &  58.99       \\ \bottomrule
\end{tabular}
\vspace{-2mm}
\end{minipage}
\end{table}

\subsection{Discussions}
\subsubsection{Model and computation complexity.} Model Complexity: The original model has $1.79$ billion (including the auto-encoder and text encoder) parameters in total, while the added adapters contain $7.49$ million parameters, leading to an increase of $0.42\%$ in memory usage. Computation Complexity: We report the peak GPU VRAM and the time required for outpainting a target video from $512 \times 512$ to $512 \times 1024$ with 16 frames at two stages in Table~\ref{tab:computationcomplexity}. For longer videos, as described in Section~\ref{sec:longvideooutpainting}, instead of processing the long video as a whole, we adapt only to short video clips sampled from the long video. This approach does not require additional time or GPU VRAM during the input-specific adaptation phase. Additionally, with temporal co-denoising~\cite{wang2023gen}, the GPU VRAM usage remains the same as that for short video during the pattern-aware outpainting phase, while the required time increases linearly with the video length.

\noindent \textbf{User study.} We conducted a user study between \ours{} and M3DDM, utilizing the DAVIS dataset with a horizontal mask of 0.66 as source videos. Preferences were collected from 10 volunteers, each evaluating 50 randomly selected sets of results based on visual quality (such as clarity, color fidelity, and texture detail) and realism (including motion consistency, object continuity, and integration with the background). Table~\ref{tab:ablation2} demonstrates that the outputs from MOTIA are preferred over those from M3DDM  in both visual quality and realism.

\begin{table}[t]
\vspace{-1mm}
\centering
\begin{minipage}[t]{0.45\textwidth}
\centering
\caption{\textbf{Computation complexity} of \ours{}.}
\vspace{-2mm}
\label{tab:computationcomplexity}
\begin{tabular}{@{}lcc@{}}
\toprule
Phase    & VARM$\downarrow$ & Time$\downarrow$\\ \midrule
Input-Specific Adapt    &    12.70 GB   & 309 Seconds     \\
Pattern-Aware Outpaint      &    5.80 GB   & 58 Seconds      \\ \midrule
\ours{}~(In total)   &     12.70 GB   &   367 Seconds \\ \bottomrule
\end{tabular}
\end{minipage}
\hfill
\begin{minipage}[t]{0.45\textwidth}
\centering
\caption{\textbf{User study} comparison between M3DDM and \ours{}.}
\label{tab:userstudy}
\vspace{-2mm}
\begin{tabular}{@{}lcc@{}}
\toprule
Method   & Visual-Quality & Realism \\ \midrule
M3DDM & 27.4\% & 42.8\% \\ \midrule
\ours{} & 72.6\% & 57.2\%\\ \bottomrule
\end{tabular}
\vspace{-2mm}
\end{minipage}
\vspace{-6mm}
\end{table}
\noindent \textbf{Why \ours{} outperforms~(Why previous methods fail).} 1) \textbf{Flexibility.} Current video diffusion models are mostly trained with fixed resolution and length, lacking the ability to tackle videos with various aspect ratios and lengths. In contrast, the adaptation phase of \ours{} allows the model to better capture the size, length, and style distribution of the source video, greatly narrowing the gap between pretrained weights and the source video. 2) \textbf{Ability for capturing intrinsic patterns from source video.} A crucial point for successful outpainting is the predicted score of diffusion models should be well-compatible with the original known regions of the source video. To achieve this, the model should effectively extract useful information from the source video for denoising. For instance, M3DDM concatenates local frames of source video at the input layers and incorporates the global frames through the cross-attention mechanism after passing light encoders. However, the information might not be properly handled especially for out-domain inputs, thus leading to outpainting failure. Instead, by conducting input-specific adaptation on the source video, the model can effectively capture the data-specific patterns in the source videos through gradient. Through this, \ours{} better leverage the data-specific patterns of the source video and image/video generative prior for outpainting. We hope this work inspires following research to exploit more from the source video itself instead of purely relying on the generative prior from intensive training on videos.

\vspace{-4mm}
\section{Conclusion}
\vspace{-3mm}
We present \ours{}, an innovative advancement in video outpainting. \ours{} relies on a combination of input-specific adaptation for capturing inner video patterns and pattern-aware outpainting to generalize these patterns for actual outpainting. Extensive experiments validate the effectiveness. 

\noindent\textbf{Limitations:} \ours{} requires learning necessary patterns from the source video, when the source video contains little information, it poses a significant challenge for \ours{} to effectively outpainting it.

\clearpage

\bibliographystyle{splncs04}
\bibliography{main}

\appendix
\clearpage
\setcounter{page}{1}

\chapter*{Supplementary Material \\MOTIA: Mastering Video Outpainting \\ through Input-Specific Adaptation}

\renewcommand{\thesection}{\Roman{section}}
\section{Detailed Implementation}
\vspace{-1mm}
\subsubsection{Model architecture.}
MOTIA marries generative priors with pretrained models for fast adaptation and better generalization. Specifically, the basic components of MOTIA in model architecture aspect are:
\begin{itemize}[label=\textbullet]
    \item Variational autoencoder~\cite{rombach2022high}. The autoencoder consists of an encoder and decoder, with the encoder mapping the original video frames into latent space and the decoder decoding the video frames with latent codes.
    \item CLIP text encoder~\cite{clip}. CLIP is trained on vast text-image pairs, enabling its text encoder to contain meaningful and rich information for controlling image generation.  
    \item U-Net~\cite{rombach2022high}. We apply Stable Diffusion v1-5 as our fundamental denoisier. The U-Net is conditioned on text embeddings of CLIP through cross-attention. To make it applicable to the 3D features of videos, we inflate the 2D convolutions and 2D group normalizations within it into pseudo-3D convolutions and 3D group normalizations.
    \item Temporal module~\cite{guo2023animatediff}. To equip the model with additional temporal priors, we initialize additional temporal attention layers with vanilla transformer architectures pretrained on large-scale text video datasets. Note that, we have shown that directly applying this temporal prior for video outpainting leads to poor results without our proposed input-specific adaptation process.
    \item LoRA~\cite{hu2021lora}. LoRA is proposed for the efficient fine-tuning of large models. It has been widely used in various diffusion-based applications, including video editing and manipulation. Therefore, we also choose LoRA as the basic learning component. Additionally, unlike previous works directly inserting the trained LoRA, we propose an effective strategy that adjusts the insertion weight of LoRA according to the spatial position of the given feature, achieving better balance in the learned patterns and generative priors of the pretrained model.
    \item ControlNet~\cite{zhang2023adding}. ControlNet works as a plug-and-play module for Stable Diffusion, allowing it to accept additional input for better controlling the denoising results.  We apply a ControlNet pretrained on Image Inpainting tasks, accepting the masked image to instruct the whole denoising process. 
    \item Blip~\cite{li2022blip}. Note that our method is built upon Stable Diffusion, which is a conditional denoiser, requiring appropriate text conditions to achieve good results. We apply Blip to automatically provide the captions to avoid man-made influence.  
\end{itemize}

\subsubsection{Pseudo algorithm code.}
The MOTIA framework operates in a two-fold manner: the input-specific adaptation hones the model's ability to capture the essential content and motion patterns from the source video, while the pattern-aware outpainting generalizes the captured patterns to creatively expand the video's horizon. The overall pipelines for input-specific adaptation and pattern-aware outpainting are shown in Algorithm~\ref{algo:intrinsicadaptation} and Algorithm~\ref{algo:extrinsicrendering}.

\begin{algorithm*}[t]
\caption{Input-Specific Adaptation in MOTIA.}\label{algo:intrinsicadaptation}
\begin{algorithmic}[1]

\Require Source video $\vv$
\State Initialize Stable Diffusion (SD), ControlNet (C), Temporal Module (T) with frozen weights
\State Initialize trainable Low-Rank Adapters (loRA)

\Function{Input-Specific Adaptation}{}
    \State Insert loRA fully into layers of SD \Comment{Full-insertion}
    \State Add loRA to the optimizer 
    \For{$i = 1$ \textbf{to} iterations}
        \State $\vv_{\text{augment}} \leftarrow$ Augment($\vv$)
        \State $t \sim$ Uniform$(1, T)$
        \State $\vepsilon \sim \mathcal{N}(\mathbf{0}, \mathbf{I})$
        \State $\vv_{\text{noisy}} \leftarrow$ AddNoise($\vv_{\text{augment}}, t, \vepsilon$) 
        \State $\vv_{\text{mask}} \leftarrow$  RandomMask($\vv_{\text{augment}}$) 
        \State Optimize gap between predicted noise and $\vepsilon$  
        \State $\gL = \left\| \vepsilon - \hat{\vepsilon}_{\bar{\vtheta_l}, \bar{\vtheta_c}, \vtheta_a}(\vv_{\text{noisy}}, \vv_{\text{masked}}, t)\right\|_2  $  \Comment{Learning through pseudo outpainting task}
        \State Gradient backpropagation
        \State Update optimizer
        \State Zero gradients of optimizer
    \EndFor
\EndFunction

\end{algorithmic}
\end{algorithm*}

\begin{algorithm*}[t]
\caption{Pattern-Aware Outpainting in MOTIA.}\label{algo:extrinsicrendering}
\begin{algorithmic}[1]

\Require Source video $\vv$
\Ensure Outpainted video $\vv_{\text{outpainted}}$

\Function{Pattern-Aware Outpainting}{}
    \State Insert loRA Spatial-awarely into layers of SD \Comment{SA-Insertion}
    \State Repeat Time $M$, and Jump Length $L$ \Comment{Hyper-parameters for noise regret}
    \State $\vv_T \sim \mathcal{N}(\mathbf{0}, \mathbf{I})$
    \State $t \gets T$
    \State $m \gets 0$
    \While{$t \neq 0$}
        \State $\boldsymbol{\epsilon} \sim \mathcal{N}(\mathbf{0}, \mathbf{I})$ if $t > 1$ else $\boldsymbol{\epsilon} = \mathbf{0}$
        \State $\vv_{t-1}^{\text{known}} \leftarrow \sqrt{\bar{\alpha}_{t-1}} \vv_0 + \sqrt{(1 - \bar{\alpha}_{t-1})} \boldsymbol{\epsilon}$
        \State $z_t \sim \mathcal{N}(\mathbf{0}, \mathbf{I})$ if $t > 1$ else $z_t = \mathbf{0}$
       \State $\boldsymbol{v}_{t-1}^{\text{unkonwn}}=\sqrt{\bar\alpha_{t-1}} \left(\frac{\boldsymbol{v}_t-\sqrt{1-\bar\alpha_t} \boldsymbol{\epsilon}_\theta\left(\boldsymbol{v}_t,t\right)}{\sqrt{\bar\alpha_t}}\right)+\sqrt{1-\bar\alpha_{t-1}-\sigma_t^2} \cdot \boldsymbol\epsilon_\theta\left(\boldsymbol{x}_t,t\right) +\sigma_t z_{t}$    
    \State $\vv_{t-1} \leftarrow \vm \odot \vv_{t-1}^{\text{known}} + (1 - \vm) \odot \vv_{t-1}^{\text{unknown}}$
        \If {$(T-t+1) \mod L = 0$}
            \If {$m < M$}
            \State $\vv_{t+L-1} \sim \mathcal{N} \left(\sqrt{\prod_{i=t}^{t+L-1}\alpha_{i}}\vv_{t-1}, \sqrt{1-\prod_{i=t}^{t+L-1}\alpha_{i}}\mathbf{I}\right)$ \Comment{Noise regret}
            \State $t \gets t + L - 1$
            \State $m \gets m + 1$
            \Else
                \State $m \gets 0$
            \EndIf
        \EndIf
    \EndWhile
    \State $\vv_\text{outpainted} = \vv_0$
    \State \Return $\vv_\text{outpainted}$
\EndFunction

\end{algorithmic}
\end{algorithm*}

\section{Benchmark Details}
The quantitative metric evaluation of MOTIA is mostly based
on DAVIS~\cite{davis} and YouTube-VOS~\cite{vos}. The DAVIS (Densely Annotated Video Segmentation) dataset is pivotal for video object segmentation research. DAVIS 2016 contains 50 videos (30 for training, 20 for testing), each featuring a single instance annotation per frame. DAVIS 2017 expands this scope with 150 videos in total (60 for training, 30 for validation, 60 for testing), annotating multiple instances per video. This dataset supports semi-supervised and unsupervised tasks, differing in the level of human input during testing. The YouTube-VOS dataset, designed for Video Object Segmentation (VOS), is a substantial benchmark with over 4,000 high-resolution YouTube videos, totaling over 340 minutes. It supports multiple VOS tasks, including semi-supervised and unsupervised video object segmentation. In our study, frames from these videos are used as inputs, cropped on the sides, without annotated foreground masks. Though designed for segmentation, these datasets are widely used to evaluate the performance of video outpainting and inpainting. MOTIA achieves superior performance compared to previous state-of-the-art methods~\cite{fan2023hierarchical,dehan,liew2023magicedit}.

\section{Additional Results}
\label{sec:additionalresults}

\begin{figure*}[htp]
\centering
\foreach \i in {0,...,63} {
    \begin{subfigure}{.075\textwidth}
        \centering
        \includegraphics[width=\linewidth]{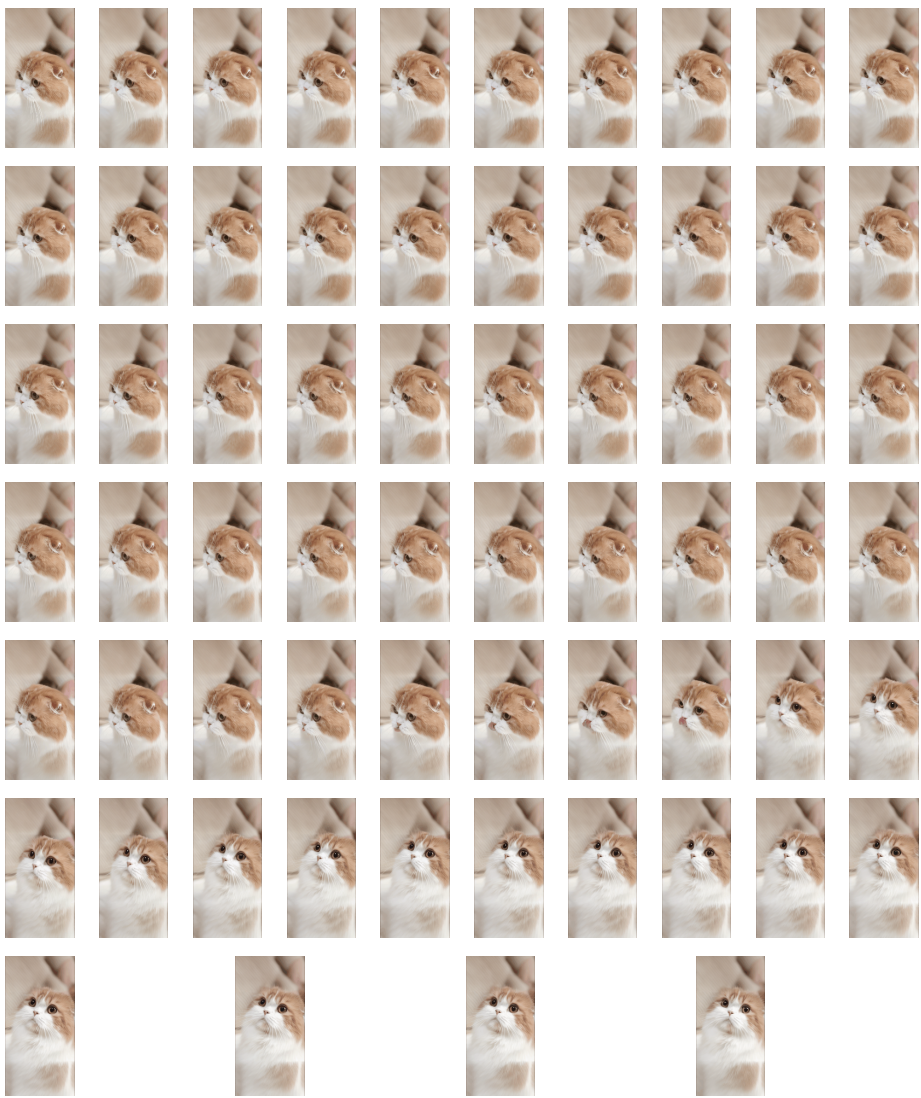}
    \end{subfigure}\hfill
    \ifnum \i=9 \par\medskip\fi
    \ifnum\i=19 \par\medskip\fi  
    \ifnum\i=29 \par\medskip\fi
    \ifnum\i=39 \par\medskip\fi
    \ifnum\i=49 \par\medskip\fi 
    \ifnum\i=59 \par\medskip\fi 
}
\caption{Results of MOTIA on long video outpainting, from $256\times 256$ to $512\times 256$.}\label{fig:long1}
\end{figure*}

\begin{figure*}[htp]
\centering
\foreach \i in {0,...,63} {
    \begin{subfigure}{.16\textwidth}
        \centering
        \includegraphics[width=\linewidth]{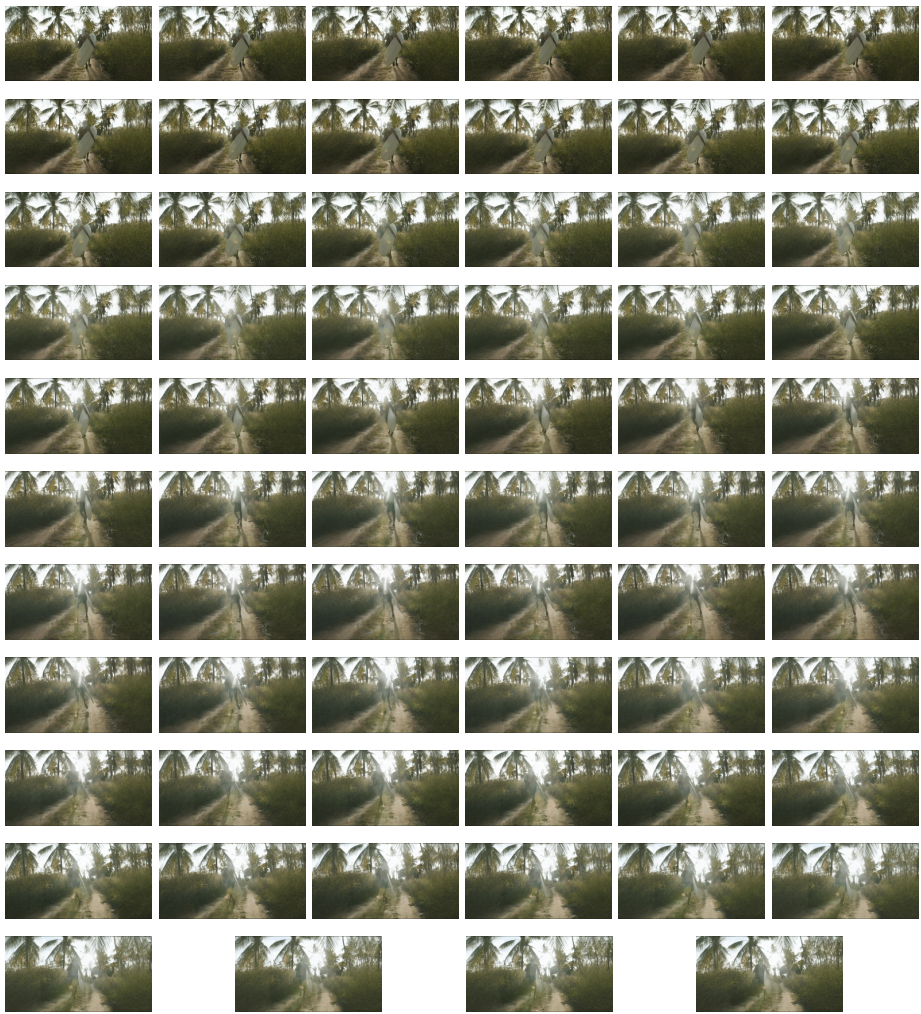}
    \end{subfigure}\hfill
    \ifnum\i=5 \par\medskip\fi
    \ifnum\i=11 \par\medskip\fi  
    \ifnum\i=17 \par\medskip\fi
    \ifnum\i=23 \par\medskip\fi
    \ifnum\i=29 \par\medskip\fi 
    \ifnum\i=35 \par\medskip\fi 
    \ifnum\i=41 \par\medskip\fi 
    \ifnum\i=47 \par\medskip\fi 
    \ifnum\i=53 \par\medskip\fi 
    \ifnum\i=59 \par\medskip\fi 
}
\caption{Results of MOTIA on long video outpainting, from $256\times 256$ to $256\times 512$.}\label{fig:long2}
\end{figure*}

\begin{figure*}[!t]
    \centering
    \begin{subfigure}[b]{0.49\textwidth}
        \includegraphics[width=\textwidth]{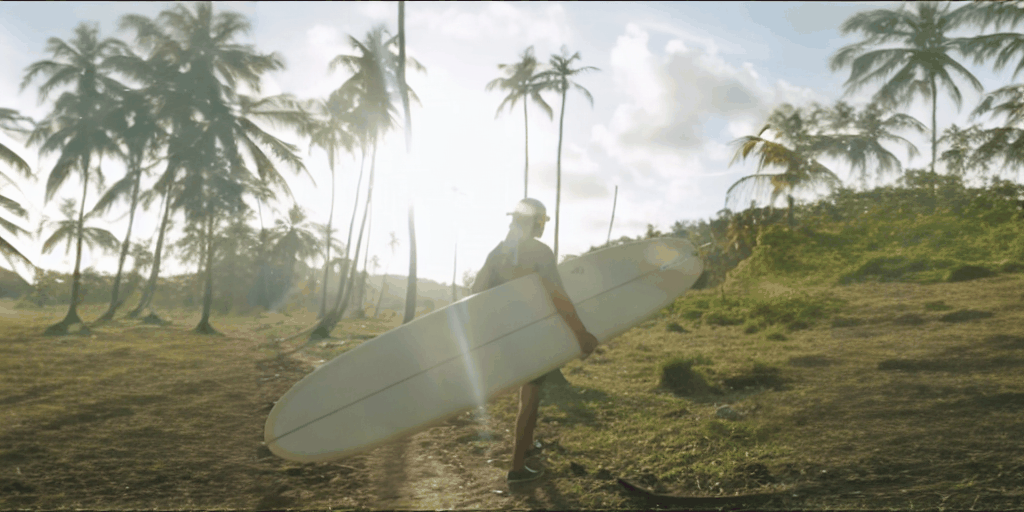}
    \end{subfigure}
    \hfill 
    \begin{subfigure}[b]{0.49\textwidth}
        \includegraphics[width=\textwidth]{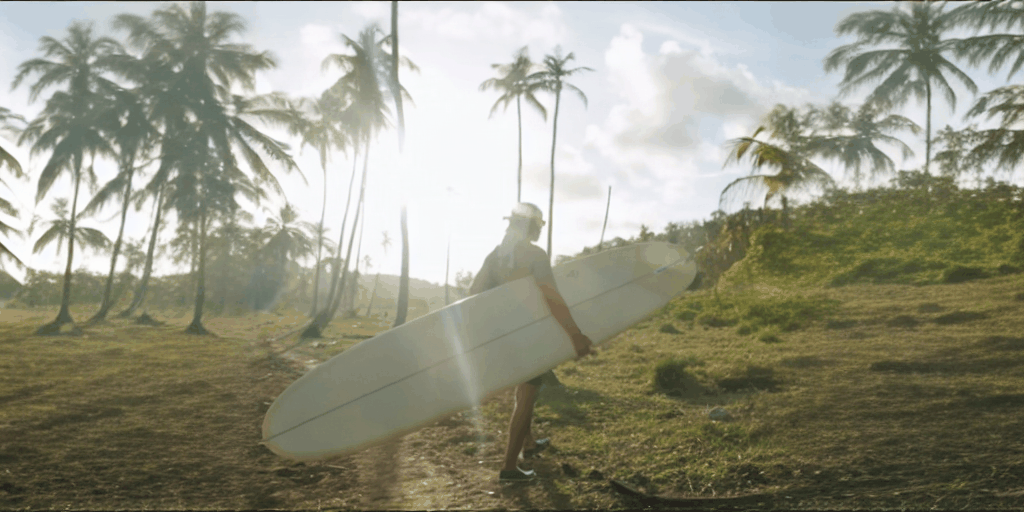}
    \end{subfigure}
    \hfill 
    \vspace{1cm}
    \begin{subfigure}[b]{0.49\textwidth}
        \includegraphics[width=\textwidth]{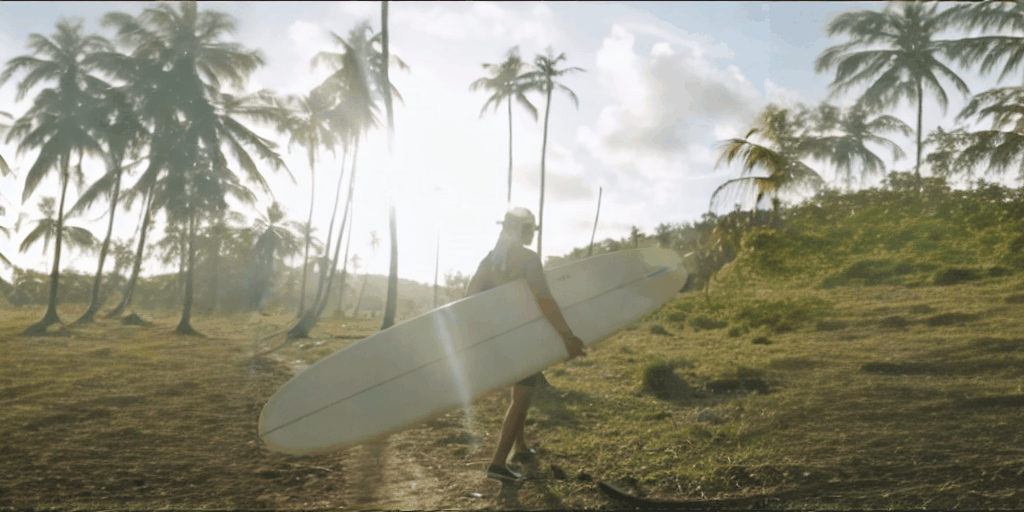}
    \end{subfigure}
    \hfill 
    \begin{subfigure}[b]{0.49\textwidth}
        \includegraphics[width=\textwidth]{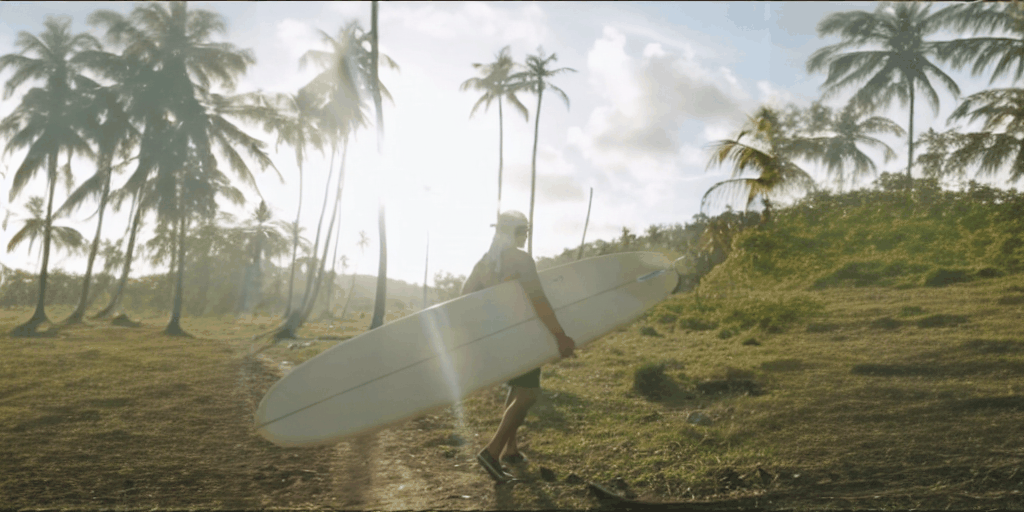}
    \end{subfigure}
    \hfill 
    \vspace{1cm}
    \begin{subfigure}[b]{0.49\textwidth}
        \includegraphics[width=\textwidth]{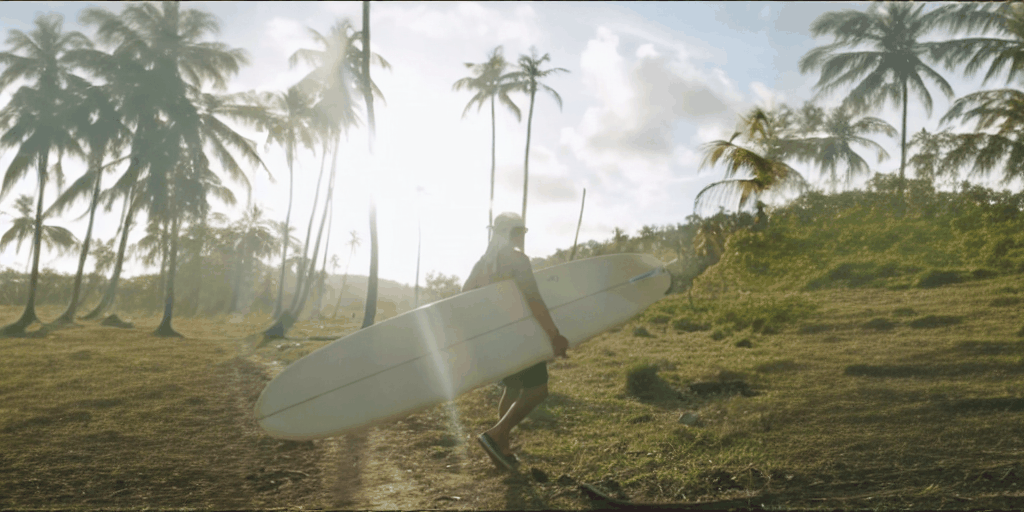}
    \end{subfigure}
    \hfill 
    \begin{subfigure}[b]{0.49\textwidth}
        \includegraphics[width=\textwidth]{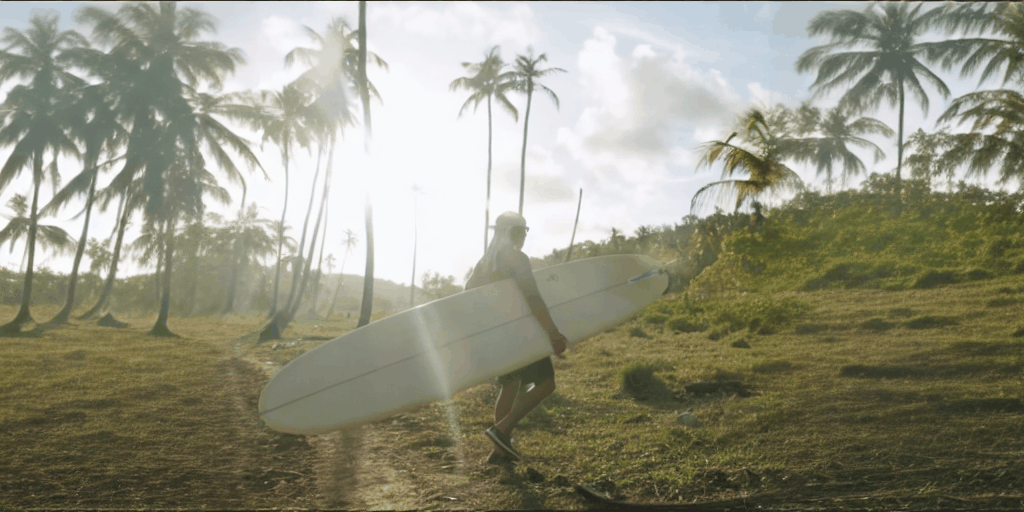}
    \end{subfigure}
    \hfill 
    \vspace{1cm}
    \begin{subfigure}[b]{0.49\textwidth}
        \includegraphics[width=\textwidth]{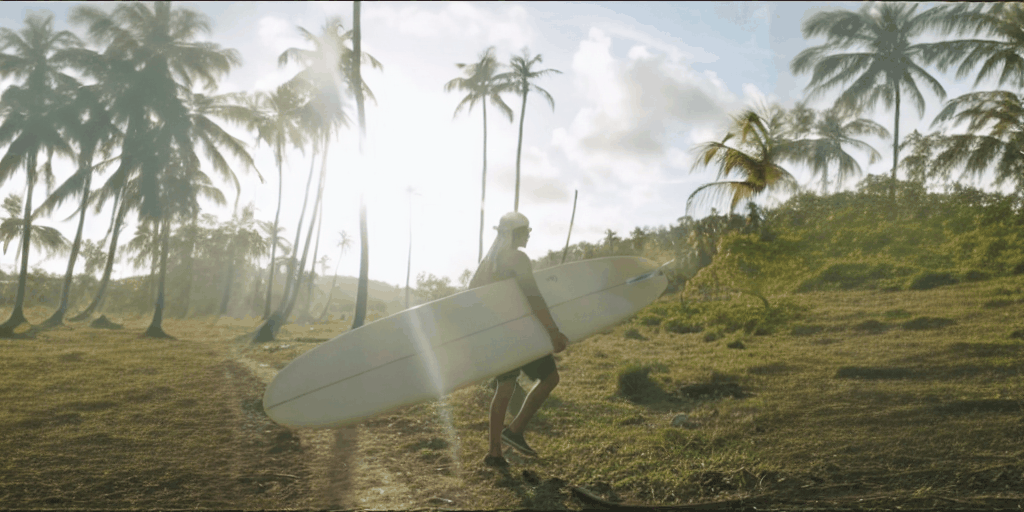}
    \end{subfigure}
    \hfill 
    \begin{subfigure}[b]{0.49\textwidth}
        \includegraphics[width=\textwidth]{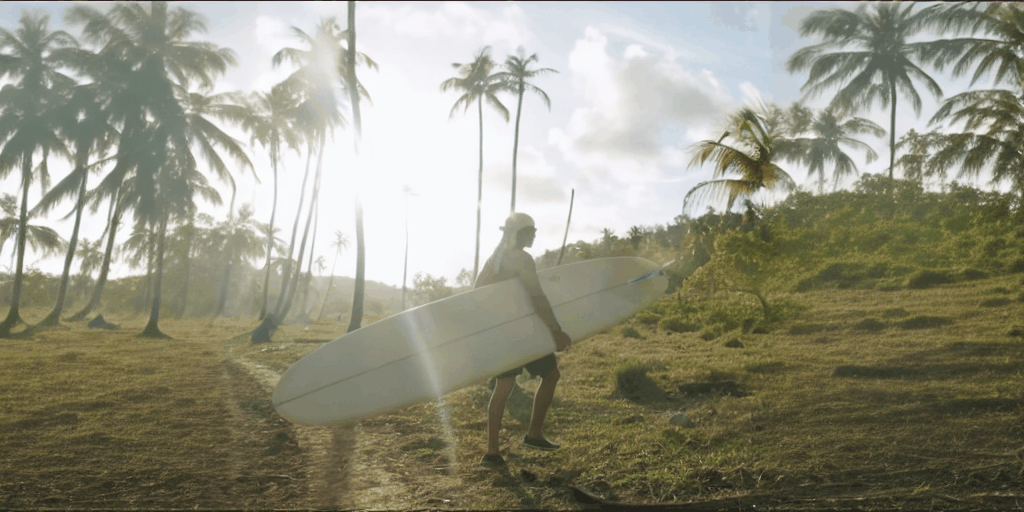}
    \end{subfigure}
    \caption{Results of MOTIA on high resolution video outpainting, from $512\times 512$ to $512\times 1024$.}
    \label{fig:addition1}
\end{figure*}

\begin{figure*}[!t]
    \centering
    \begin{subfigure}[b]{0.49\textwidth}
        \includegraphics[width=\textwidth]{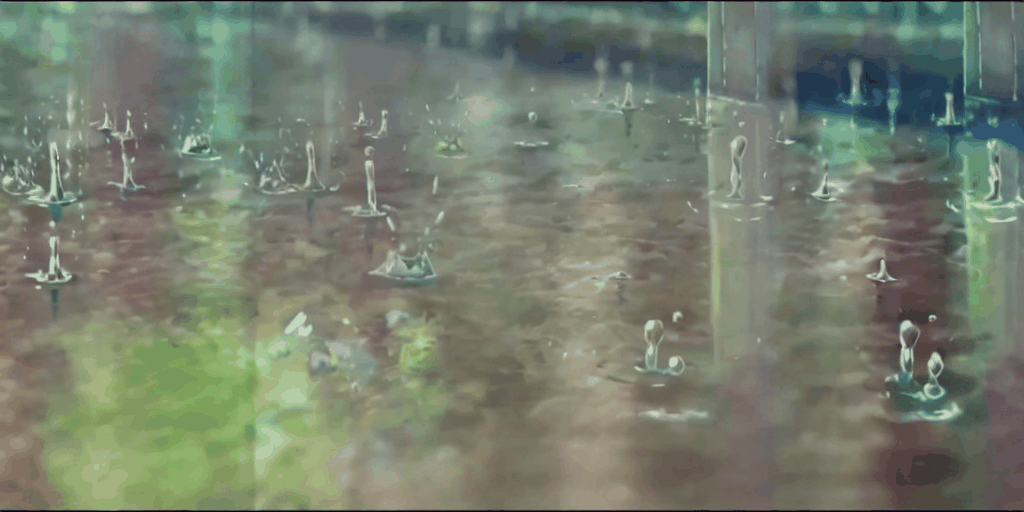}
    \end{subfigure}
    \hfill 
    \begin{subfigure}[b]{0.49\textwidth}
        \includegraphics[width=\textwidth]{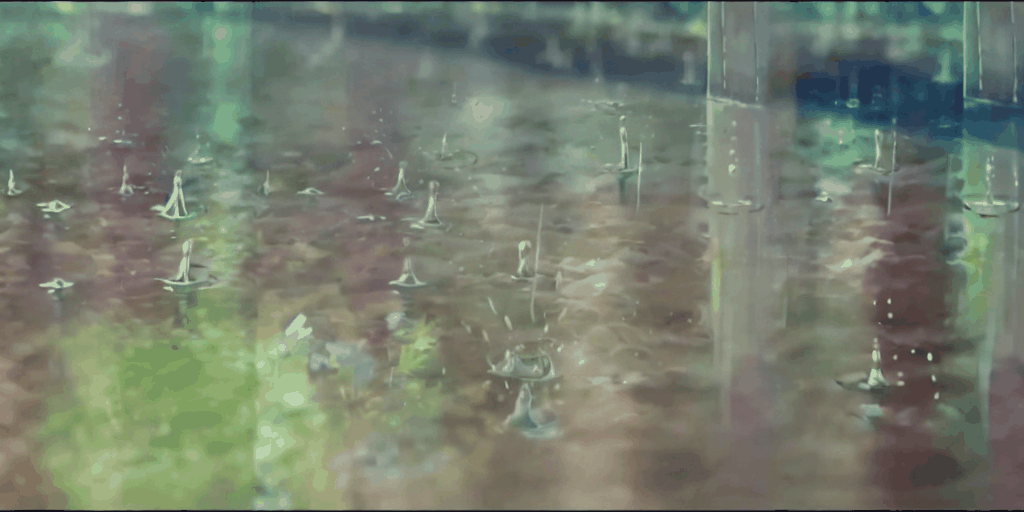}
    \end{subfigure}
    \hfill 
    \vspace{1cm}
    \begin{subfigure}[b]{0.49\textwidth}
        \includegraphics[width=\textwidth]{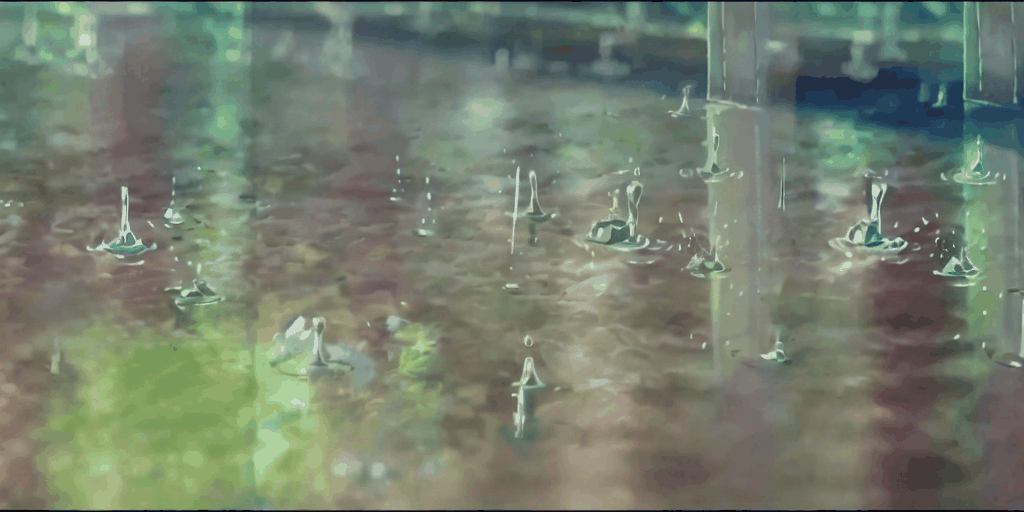}
    \end{subfigure}
    \hfill
    \begin{subfigure}[b]{0.49\textwidth}
        \includegraphics[width=\textwidth]{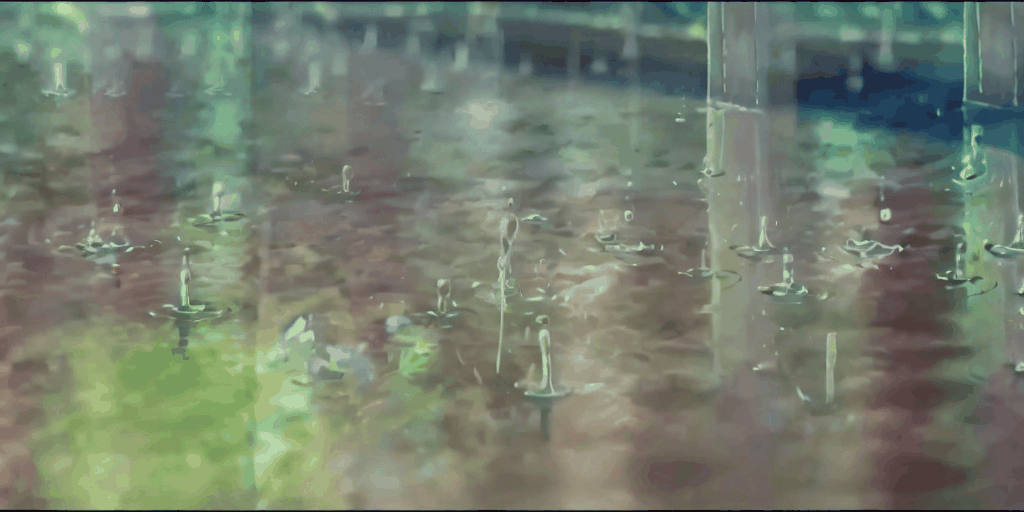}
    \end{subfigure}
    \hfill 
    \vspace{1cm}
    \begin{subfigure}[b]{0.49\textwidth}
        \includegraphics[width=\textwidth]{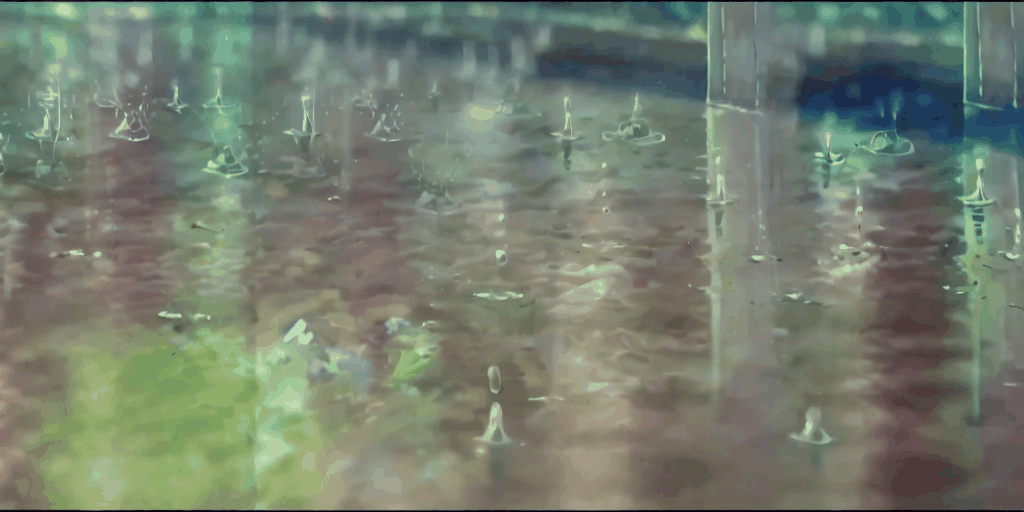}
    \end{subfigure}
    \hfill 
    \begin{subfigure}[b]{0.49\textwidth}
        \includegraphics[width=\textwidth]{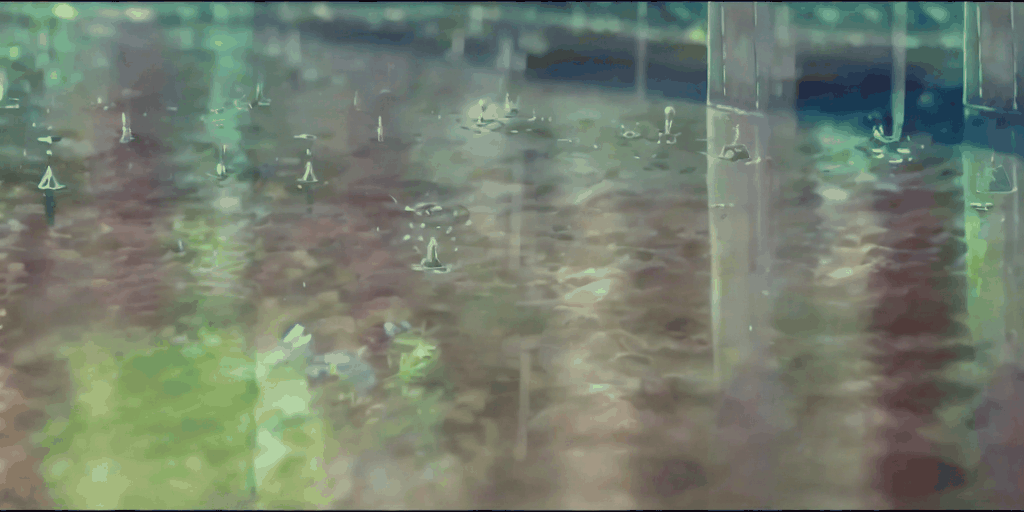}
    \end{subfigure}
    \hfill 
    \vspace{1cm}
    \begin{subfigure}[b]{0.49\textwidth}
        \includegraphics[width=\textwidth]{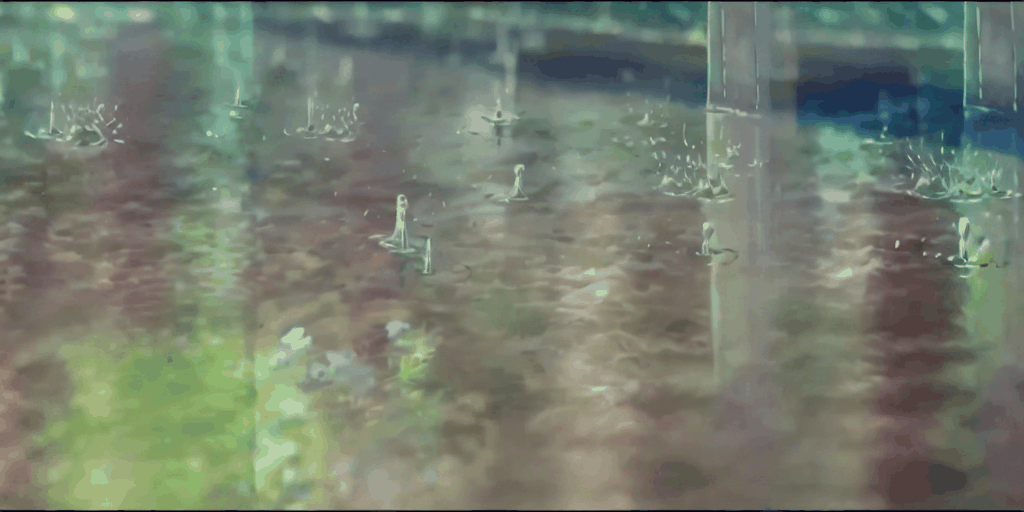}
    \end{subfigure}
    \hfill 
    \begin{subfigure}[b]{0.49\textwidth}
        \includegraphics[width=\textwidth]{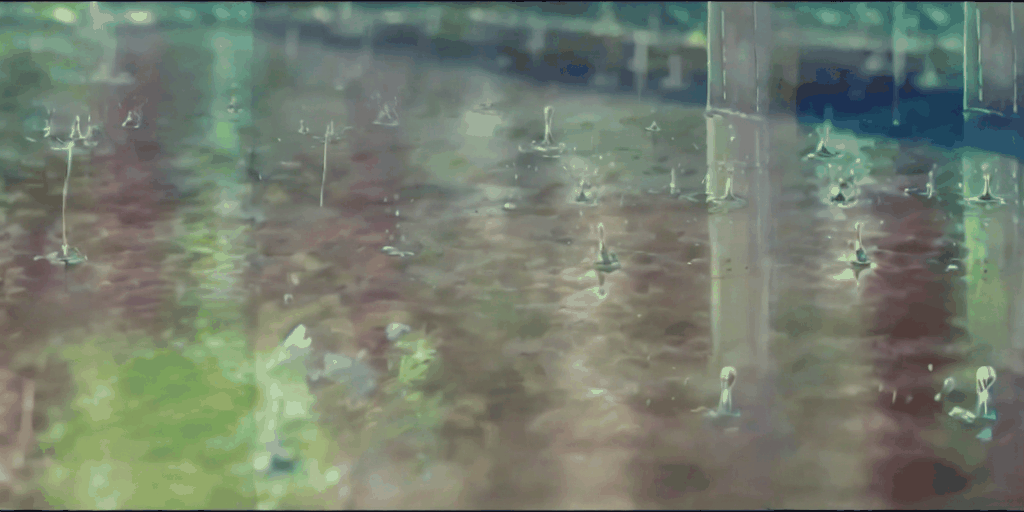}
    \end{subfigure}
    \caption{Results of MOTIA on high resolution video outpainting, from $512\times 512$ to $512\times 1024$.}        \label{fig:addition2}
\end{figure*}

\begin{figure*}[!t]
    \centering
    \begin{subfigure}[b]{0.49\textwidth}
        \includegraphics[width=\textwidth]{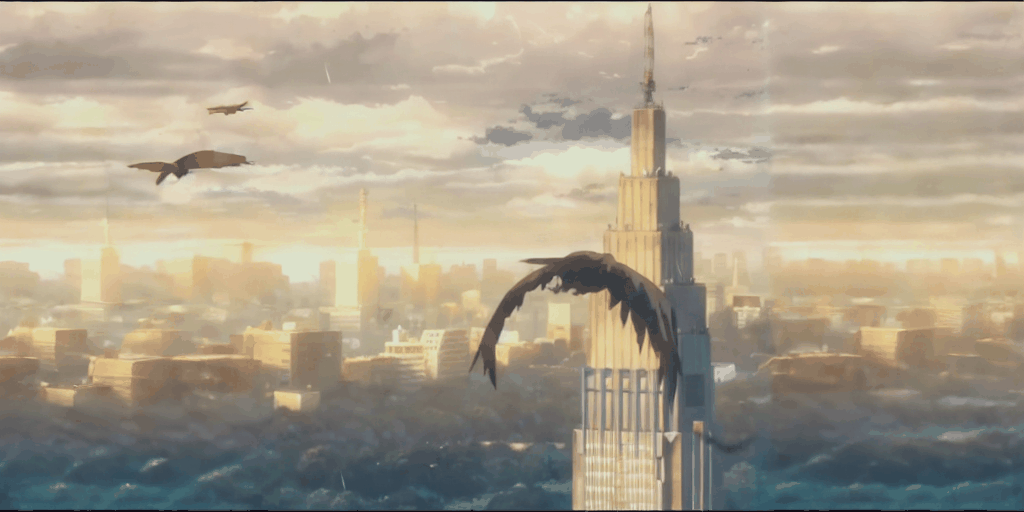}
    \end{subfigure}
    \hfill 
    \begin{subfigure}[b]{0.49\textwidth}
        \includegraphics[width=\textwidth]{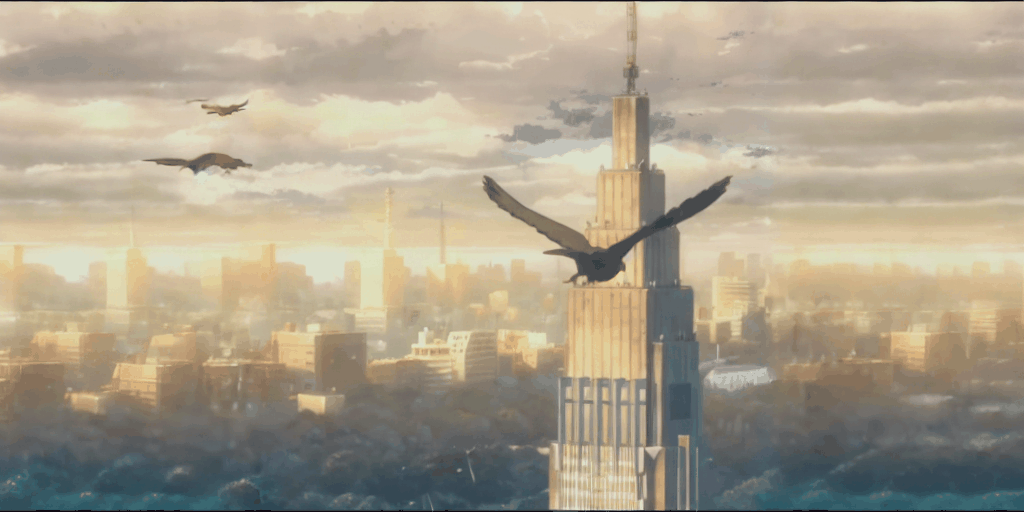}
    \end{subfigure}
    \hfill 
    \vspace{1cm}
    \begin{subfigure}[b]{0.49\textwidth}
        \includegraphics[width=\textwidth]{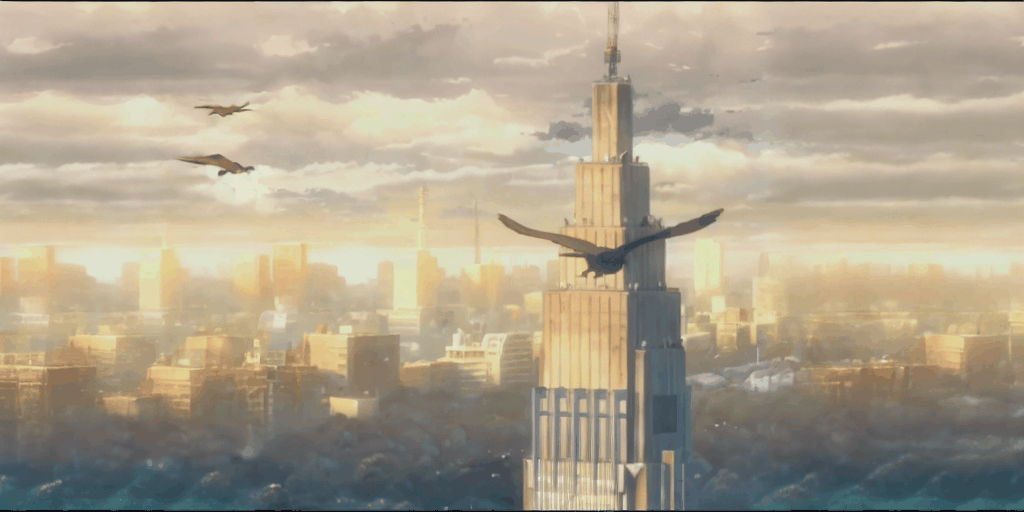}
    \end{subfigure}
    \hfill 
    \begin{subfigure}[b]{0.49\textwidth}
        \includegraphics[width=\textwidth]{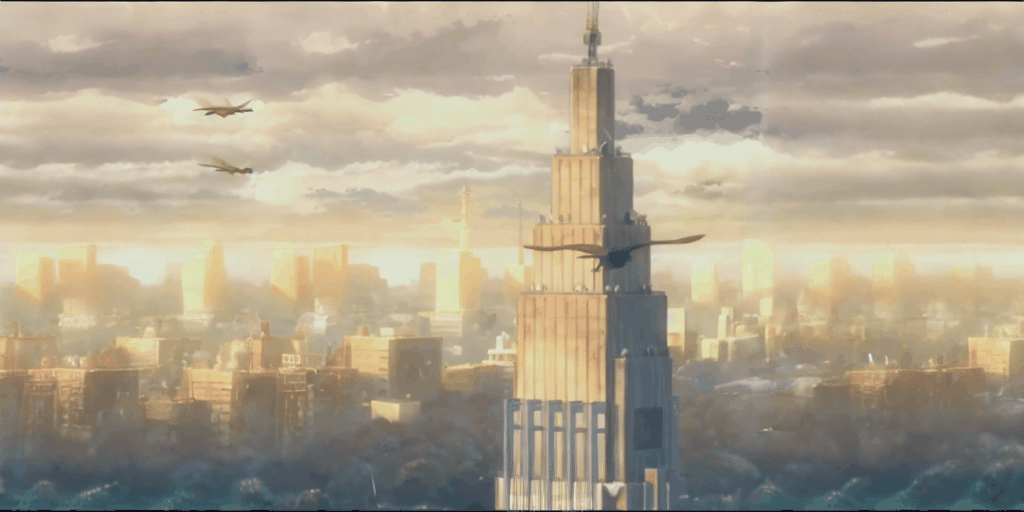}
    \end{subfigure}
    \hfill
    \vspace{1cm}
    \begin{subfigure}[b]{0.49\textwidth}
        \includegraphics[width=\textwidth]{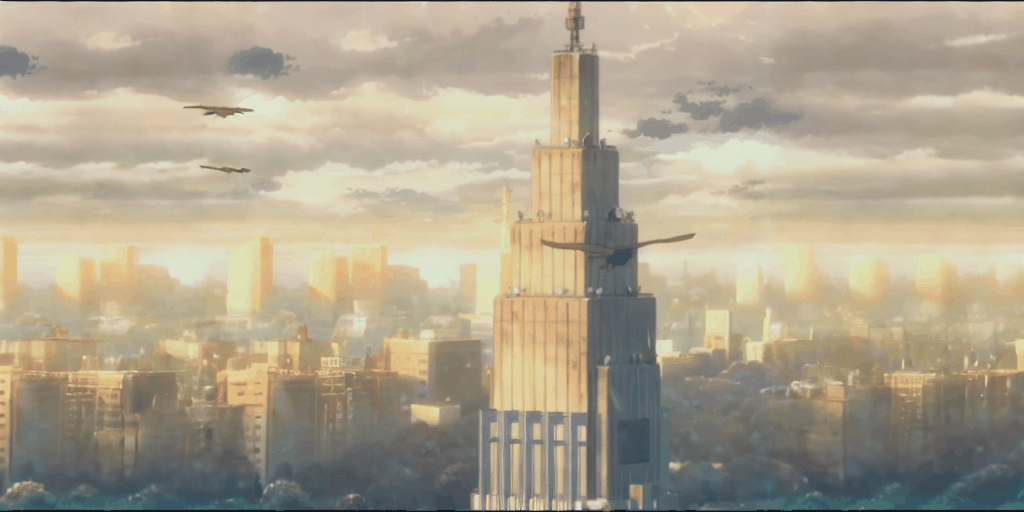}
    \end{subfigure}
    \hfill 
    \begin{subfigure}[b]{0.49\textwidth}
        \includegraphics[width=\textwidth]{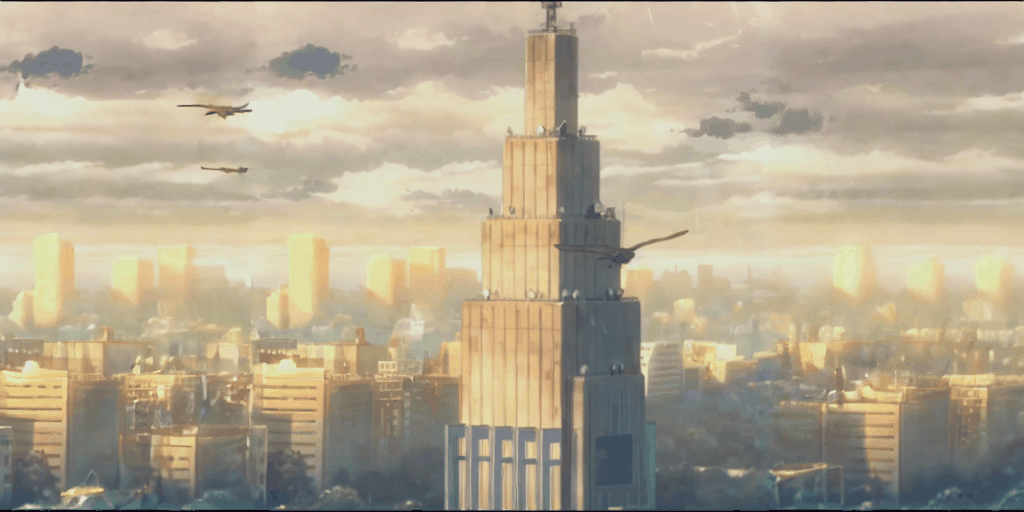}
    \end{subfigure}
    \hfill 
    \vspace{1cm}
    \begin{subfigure}[b]{0.49\textwidth}
        \includegraphics[width=\textwidth]{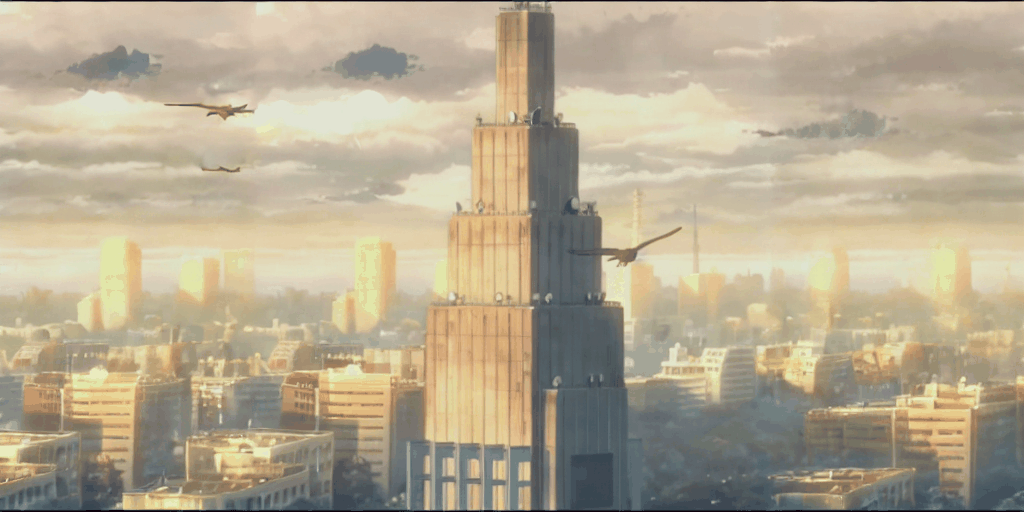}
    \end{subfigure}
    \hfill 
    \begin{subfigure}[b]{0.49\textwidth}
        \includegraphics[width=\textwidth]{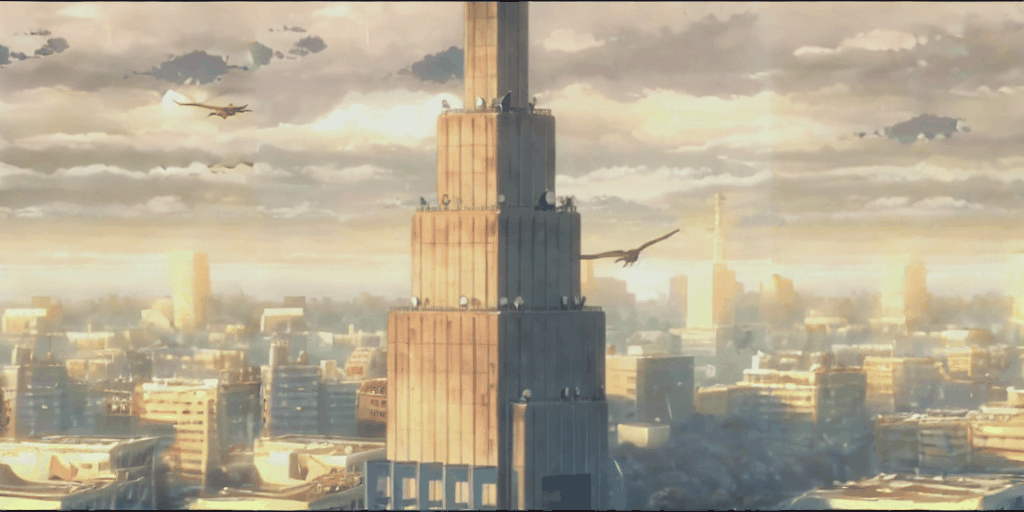}
    \end{subfigure}
    \caption{Results of MOTIA on high resolution video outpainting, from $512\times 512$ to $512\times 1024$.}        \label{fig:addition3}
\end{figure*}

We report additional results outpainted by MOTIA. Fig.~\ref{fig:long1} and Fig.~\ref{fig:long2}  show the longer videos~(8 seconds compared to baseline 2 seconds) outpainted by MOTIA. Fig.~\ref{fig:addition1}, Fig.~\ref{fig:addition2}, Fig.~\ref{fig:addition3} show high resolution videos outpainted by MOTIA. 

\section{Demo  Video}
We provide a demo video, which can be viewed on the anonymous project page or the supplementary video file,  showing:

\noindent\textbf{Outpainting results.} The results cover videos with various subjects and styles in different resolutions and video lengths, showing the versatile applicability of MOTIA.

\noindent\textbf{Baseline comparison.}
We compare the outpainting results of MOTIA and previous methods in different settings. The results show that MOTIA surpasses previous methods in visual quality, frame consistency, and the harmony of the outpaint scenes in videos.

\end{document}